\documentclass{article}

\usepackage{times}
\usepackage{graphicx} 
\usepackage{subfigure}

\usepackage{natbib}

\usepackage{algorithm}
\usepackage{algorithmic}
\usepackage[algo2e,linesnumbered,lined,boxed,vlined]{algorithm2e}

\usepackage{amsfonts}
\usepackage{amssymb}

\usepackage{hyperref}

\usepackage[accepted]{icml2016}

\usepackage{comment}
\usepackage{amsthm}
\usepackage{mathtools}
\usepackage{tabularx}
\usepackage{multirow}

\def\b{\ensuremath\boldsymbol}

\usepackage{lastpage}
\usepackage{fancyhdr}
\usepackage{url}
\icmltitlerunning{Fisher and Kernel Fisher Discriminant Analysis: Tutorial}

\begin{document}

\twocolumn[
\icmltitle{Fisher and Kernel Fisher Discriminant Analysis: Tutorial}

\icmlauthor{Benyamin Ghojogh}{bghojogh@uwaterloo.ca}
\icmladdress{Department of Electrical and Computer Engineering, 
\\Machine Learning Laboratory, University of Waterloo, Waterloo, ON, Canada}
\icmlauthor{Fakhri Karray}{karray@uwaterloo.ca}
\icmladdress{Department of Electrical and Computer Engineering, 
\\Centre for Pattern Analysis and Machine Intelligence, University of Waterloo, Waterloo, ON, Canada}
\icmlauthor{Mark Crowley}{mcrowley@uwaterloo.ca}
\icmladdress{Department of Electrical and Computer Engineering, 
\\Machine Learning Laboratory, University of Waterloo, Waterloo, ON, Canada}

\icmlkeywords{Tutorial, Principal Component Analysis}

\vskip 0.3in
]

\begin{abstract}
This is a detailed tutorial paper which explains the Fisher discriminant Analysis (FDA) and kernel FDA. We start with projection and reconstruction. Then, one- and multi-dimensional FDA subspaces are covered. Scatters in two- and then multi-classes are explained in FDA. Then, we discuss on the rank of the scatters and the dimensionality of the subspace. A real-life example is also provided for interpreting FDA. Then, possible singularity of the scatter is discussed to introduce robust FDA. PCA and FDA directions are also compared. We also prove that FDA and linear discriminant analysis are equivalent. Fisher forest is also introduced as an ensemble of fisher subspaces useful for handling data with different features and dimensionality. Afterwards, kernel FDA is explained for both one- and multi-dimensional subspaces with both two- and multi-classes. Finally, some simulations are performed on AT\&T face dataset to illustrate FDA and compare it with PCA. 
\end{abstract}

\section{Introduction}

Assume we have a dataset of \textit{instances} or \textit{data points} $\{(\b{x}_i, \b{y}_i)\}_{i=1}^n$ with sample size $n$ and dimensionality $\b{x}_i  \in \mathbb{R}^d$ and $\b{y}_i \in \mathbb{R}^\ell$. 
The $\{\b{x}_i\}_{i=1}^n$ are the input data to the model and the $\{\b{y}_i\}_{i=1}^n$ are the observations (labels).
We define $\mathbb{R}^{d \times n} \ni \b{X} := [\b{x}_1, \dots, \b{x}_n]$ and $\mathbb{R}^{\ell \times n} \ni \b{Y} := [\b{y}_1, \dots, \b{y}_n]$.
We can also have an out-of-sample data point, $\b{x}_t \in \mathbb{R}^d$, which is not in the training set. If there are $n_t$ out-of-sample data points, $\{\b{x}_{t,i}\}_1^{n_t}$, we define $\mathbb{R}^{d \times n_t} \ni \b{X}_t := [\b{x}_{t,1}, \dots, \b{x}_{t,n_t}]$.
Usually, the data points exist on a subspace or sub-manifold. Subspace or manifold learning tries to learn this sub-manifold \cite{ghojogh2019feature}.

Here, we consider the case where the observations  $\{\b{y}_i\}_{i=1}^n$ come from a discrete set so that the task is \textit{classification}. 
Assume the dataset consists of $c$ classes, $\{\b{x}_i^{(1)}\}_{i=1}^{n_1}, \dots, \{\b{x}_i^{(c)}\}_{i=1}^{n_c}$ where $n_j$ denotes the sample size (cardinality) of the $j$-th class. 

We want to find a subspace (or sub-manifold) which separates the classes as much as possible while the data also become as spread as possible. 
Fisher Discriminant Analysis (FDA) \cite{friedman2001elements} pursues this goal. It was first proposed in \cite{fisher1936use} by Sir. Ronald Aylmer Fisher (1890 -- 1962) who was a genius in statistics. He proposed many important concepts in the modern statistics, such as variance \cite{fisher1919xv}, FDA \cite{fisher1936use}, Fisher information \cite{frieden2004science}, Analysis of Variance (ANOVA) \cite{fisher1992statistical}, etc. The paper \cite{fisher1936use}, which proposed FDA, was the first paper introducing the well-known Iris flower dataset. 
Note that Fisher's work was mostly concentrating on the statistics in the area of genetics. 
Much of his work was about variance making no wonder for us why FDA is all about variance and scatters.

Kernel FDA \cite{mika1999fisher,mika2000invariant} performs the goal of FDA in the feature space. 
The FDA and kernel FDA have had many different applications. 
Some examples for applications of FDA are facial recognition (Fisherfaces) \cite{belhumeur1997eigenfaces,etemad1997discriminant,zhao1999subspace}, action recognition (Fisherposes) \cite{ghojogh2017fisherposes,mokari2018recognizing}, and gesture recognition \cite{samadani2013discriminative}.
Some examples for applications of kernel FDA are facial recognition (kernel Fisherfaces) \cite{yang2002kernel,liu2004improving} and palmprint recognition \cite{wang2006kernel}.

In the literature, sometimes, FDA is referred to as Linear Discriminant Analysis (LDA) or Fisher LDA (FLDA). This is because FDA and LDA \cite{ghojogh2019linear} are equivalent although LDA is a classification method and not a subspace learning algorithm.
In this paper, we will prove why they are equivalent. 

\section{Projection Formulation}

\subsection{Projection}

Assume we have a data point $\b{x} \in \mathbb{R}^d$. We want to project this data point onto the vector space spanned by $p$ vectors $\{\b{u}_1, \dots, \b{u}_p\}$ where each vector is $d$-dimensional and usually $p \ll d$. We stack these vectors column-wise in matrix $\b{U} = [\b{u}_1, \dots, \b{u}_p] \in \mathbb{R}^{d \times p}$. In other words, we want to project $\b{x}$ onto the column space of $\b{U}$, denoted by $\mathbb{C}\text{ol}(\b{U})$.

The projection of $\b{x} \in \mathbb{R}^d$ onto $\mathbb{C}\text{ol}(\b{U}) \in \mathbb{R}^p$ and then its representation in the $\mathbb{R}^d$ (its reconstruction) can be seen as a linear system of equations:
\begin{align}\label{equation_projection}
\mathbb{R}^d \ni \widehat{\b{x}} := \b{U \beta},
\end{align}
where we should find the unknown coefficients $\b{\beta} \in \mathbb{R}^p$. 

\begin{figure}[!t]
\centering
\includegraphics[width=2.2in]{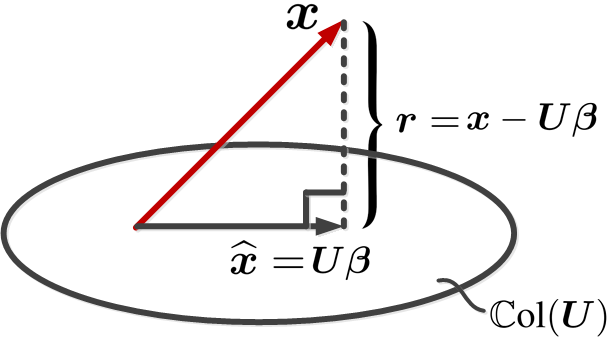}
\caption{The residual and projection onto the column space of $\b{U}$.}
\label{figure_residual_and_space}
\end{figure}

If the $\b{x}$ lies in the $\mathbb{C}\text{ol}(\b{U})$ or $\textbf{span}\{\b{u}_1, \dots, \b{u}_p\}$, this linear system has exact solution, so $\widehat{\b{x}} = \b{x} = \b{U \beta}$. However, if $\b{x}$ does not lie in this space, there is no any solution $\b{\beta}$ for $\b{x} = \b{U \beta}$ and we should solve for projection of $\b{x}$ onto $\mathbb{C}\text{ol}(\b{U})$ or $\textbf{span}\{\b{u}_1, \dots, \b{u}_p\}$ and then its reconstruction. In other words, we should solve for Eq. (\ref{equation_projection}). In this case, $\widehat{\b{x}}$ and $\b{x}$ are different and we have a residual:
\begin{align}\label{equation_residual_1}
\b{r} = \b{x} - \widehat{\b{x}} = \b{x} - \b{U \beta},
\end{align}
which we want to be small. As can be seen in Fig. \ref{figure_residual_and_space}, the smallest residual vector is orthogonal to $\mathbb{C}\text{ol}(\b{U})$; therefore:
\begin{align}
\b{x} - \b{U\beta} \perp \b{U} &\implies \b{U}^\top (\b{x} - \b{U \beta}) = 0, \nonumber \\
& \implies \b{\beta} = (\b{U}^\top \b{U})^{-1} \b{U}^\top \b{x}. \label{equation_beta}
\end{align}
It is noteworthy that the Eq. (\ref{equation_beta}) is also the formula of coefficients in linear regression \cite{friedman2001elements} where the input data are the rows of $\b{U}$ and the labels are $\b{x}$; however, our goal here is different. 

Plugging Eq. (\ref{equation_beta}) in Eq. (\ref{equation_projection}) gives us:
\begin{align*}
\widehat{\b{x}} = \b{U} (\b{U}^\top \b{U})^{-1} \b{U}^\top \b{x}.
\end{align*}
We define:
\begin{align}\label{equation_hat_matrix}
\mathbb{R}^{d \times d} \ni \b{\Pi} := \b{U} (\b{U}^\top \b{U})^{-1} \b{U}^\top,
\end{align}
as ``projection matrix'' because it projects $\b{x}$ onto $\mathbb{C}\text{ol}(\b{U})$ (and reconstructs back).
Note that $\b{\Pi}$ is also referred to as the ``hat matrix'' in the literature because it puts a hat on top of $\b{x}$.

If the vectors $\{\b{u}_1, \dots, \b{u}_p\}$ are orthonormal (the matrix $\b{U}$ is orthogonal), we have $\b{U}^\top = \b{U}^{-1}$ and thus $\b{U}^\top \b{U} = \b{I}$. Therefore, Eq. (\ref{equation_hat_matrix}) is simplified:
\begin{align}
& \b{\Pi} = \b{U} \b{U}^\top.
\end{align}
So, we have:
\begin{align}\label{equation_x_hat}
\widehat{\b{x}} = \b{\Pi}\, \b{x} = \b{U} \b{U}^\top \b{x}.
\end{align}

\subsection{Projection onto a Subspace}

In subspace learning, the projection of a vector $\b{x} \in \mathbb{R}^d$ onto the column space of $\b{U} \in \mathbb{R}^{d \times p}$ (a $p$-dimensional subspace spanned by $\{\b{u}_j\}_{j=1}^p$ where $\b{u}_j \in \mathbb{R}^d$) is defined as:
\begin{align}
&\mathbb{R}^{p} \ni \widetilde{\b{x}} := \b{U}^\top \b{x}, \label{equation_projection_training_onePoint_severalDirections} \\
&\mathbb{R}^{d} \ni \widehat{\b{x}} := \b{U}\b{U}^\top \b{x} = \b{U} \widetilde{\b{x}}, \label{equation_reconstruction_training_onePoint_severalDirections}
\end{align}
where $\widetilde{\b{x}}$ and $\widehat{\b{x}}$ denote the projection and reconstruction of $\b{x}$, respectively.

If we have $n$ data points, $\{\b{x}_i\}_{i=1}^n$, which can be stored column-wise in a matrix $\b{X} \in \mathbb{R}^{d \times n}$, the projection and reconstruction of $\b{X}$ are defined as:
\begin{align}
&\mathbb{R}^{p \times n} \ni \widetilde{\b{X}} := \b{U}^\top \b{X}, \label{equation_projection_training_SeveralPoints_severalDirections} \\
&\mathbb{R}^{d \times n} \ni \widehat{\b{X}} := \b{U}\b{U}^\top \b{X} = \b{U} \widetilde{\b{X}}, \label{equation_reconstruction_training_SeveralPoints_severalDirections}
\end{align}
respectively.

If we have an out-of-sample data point $\b{x}_t$ which was not used in calculation of $\b{U}$, the projection and reconstruction of it are defined as:
\begin{align}
&\mathbb{R}^{p} \ni \widetilde{\b{x}}_t := \b{U}^\top \b{x}_t, \label{equation_projection_outOfSample_onePoint_severalDirections} \\
&\mathbb{R}^{d} \ni \widehat{\b{x}}_t := \b{U}\b{U}^\top \b{x}_t = \b{U} \widetilde{\b{x}}_t, \label{equation_reconstruction_outOfSample_onePoint_severalDirections}
\end{align}
respectively.

In case we have $n_t$ out-of-sample data points, $\{\b{x}_{t,i}\}_{i=1}^{n_t}$, which can be stored column-wise in a matrix $\b{X}_t \in \mathbb{R}^{d \times n_t}$, the projection and reconstruction of $\b{X}_t$ are defined as:
\begin{align}
&\mathbb{R}^{p \times n_t} \ni \widetilde{\b{X}}_t := \b{U}^\top \b{X}_t, \label{equation_projection_outOfSample_SeveralPoints_severalDirections} \\
&\mathbb{R}^{d \times n_t} \ni \widehat{\b{X}}_t := \b{U}\b{U}^\top \b{X}_t = \b{U} \widetilde{\b{X}}_t, \label{equation_reconstruction_outOfSample_SeveralPoints_severalDirections}
\end{align}
respectively.

For the properties of the projection matrix $\b{U}$, refer to \cite{ghojogh2019unsupervised}.

\subsubsection{Projection onto a One-dimensional Subspace}

Considering the data $\{\b{x}_i\}_{i=1}^n$, the mean of data is:
\begin{align}
&\mathbb{R}^{d} \ni \b{\mu}_x := \frac{1}{n} \sum_{i=1}^n \b{x}_i,
\end{align}
and the centered data point $\b{x}$ is:
\begin{align}\label{equation_centered_training_onePoint}
&\mathbb{R}^{d} \ni \breve{\b{x}} := \b{x} - \b{\mu}_x.
\end{align}
The centered data $\b{X}$ is:
\begin{align}\label{equation_centered_training_SeveralPoints}
&\mathbb{R}^{d \times n} \ni \breve{\b{X}} := \b{X} - \b{\mu}_x = \b{X}\b{H},
\end{align}
where $\breve{\b{X}} = [\breve{\b{x}}_1, \dots, \breve{\b{x}}_n] \in \mathbb{R}^{d \times n}$ and $\mathbb{R}^{n \times n} \ni \b{H} := \b{I} - (1/n) \b{1}\b{1}^\top$ is the centering matrix (see Appendix A in \cite{ghojogh2019unsupervised}).

In Eq. (\ref{equation_reconstruction_training_onePoint_severalDirections}), if $p=1$, we are projecting $\b{x}$ onto only one vector $\b{u}$ and reconstruct it. If the data point is centered, the reconstruction is:
\begin{align*}
\widehat{\b{x}} = \b{u}\b{u}^\top \breve{\b{x}}.
\end{align*}
The squared length (squared $\ell_2$-norm) of this reconstructed vector is:
\begin{align}
&||\widehat{\b{x}}||_2^2 = ||\b{u}\b{u}^\top \breve{\b{x}}||_2^2 = (\b{u}\b{u}^\top \breve{\b{x}})^\top (\b{u}\b{u}^\top \breve{\b{x}}) \nonumber \\
& = \breve{\b{x}}^\top \b{u} \underbrace{\b{u}^\top \b{u}}_{1} \b{u}^\top \breve{\b{x}} \overset{(a)}{=} \breve{\b{x}}^\top \b{u}\, \b{u}^\top \breve{\b{x}} \overset{(b)}{=} \b{u}^\top \breve{\b{x}}\, \breve{\b{x}}^\top \b{u}, \label{equation_x_hat_length_squared}
\end{align}
where $(a)$ is because $\b{u}$ is a unit (normal) vector, i.e., $\b{u}^\top \b{u} = ||\b{u}||_2^2 = 1$, and $(b)$ is because $\breve{\b{x}}^\top \b{u} = \b{u}^\top \breve{\b{x}} \in \mathbb{R}$.

Suppose we have $n$ data points $\{\b{x}_i\}_{i=1}^n$ where $\{\breve{\b{x}}_i\}_{i=1}^n$ are the centered data.
The summation of the squared lengths of their projections $\{\widehat{\b{x}}_i\}_{i=1}^n$ is:
\begin{align}\label{equation_sum_projected}
\sum_{i=1}^n ||\widehat{\b{x}}_i||_2^2 \overset{(\ref{equation_x_hat_length_squared})}{=} \sum_{i=1}^n \b{u}^\top \breve{\b{x}}_i\, \breve{\b{x}}_i^\top \b{u} = \b{u}^\top \Big(\sum_{i=1}^n \breve{\b{x}}_i\, \breve{\b{x}}_i^\top\Big) \b{u}.
\end{align}
Considering $\breve{\b{X}} = [\breve{\b{x}}_1, \dots, \breve{\b{x}}_n] \in \mathbb{R}^{d \times n}$, we have:
\begin{align}
\mathbb{R}^{d \times d} \ni \b{S} &:= \sum_{i=1}^n (\b{x}_i - \b{\mu}_x)\, (\b{x}_i - \b{\mu}_x)^\top \overset{(\ref{equation_centered_training_onePoint})}{=} \sum_{i=1}^n \breve{\b{x}}_i\, \breve{\b{x}}_i^\top \nonumber \\
&= \breve{\b{X}} \breve{\b{X}}^\top \overset{(\ref{equation_centered_training_SeveralPoints})}{=} \b{X}\b{H} \b{H}\b{X}^\top, \label{equation_covariance_matrix}
\end{align}
where $\b{S}$ is called the ``covariance matrix'' or ``scatter matrix''. If the data were already centered, we would have $\b{S} = \b{X} \b{X}^\top$.

Plugging Eq. (\ref{equation_covariance_matrix}) in Eq. (\ref{equation_sum_projected}) gives us:
\begin{align}\label{equation_variance_of_projection_oneDirection1}
\sum_{i=1}^n ||\widehat{\b{x}}_i||_2^2 = \b{u}^\top \b{S} \b{u}.
\end{align}
Note that we can also say that $\b{u}^\top \b{S} \b{u}$ is the variance of the projected data onto PCA subspace. In other words, $\b{u}^\top \b{S} \b{u} = \mathbb{V}\text{ar}(\b{u}^\top \breve{\b{X}})$. This makes sense because when some non-random thing (here $\b{u}$) is multiplied to the random data (here $\breve{\b{X}}$), it will have squared (quadratic) effect on variance, and $\b{u}^\top \b{S} \b{u}$ is quadratic in $\b{u}$.

Therefore, $\b{u}^\top \b{S} \b{u}$ can be interpreted in two ways: (I) the squared length of reconstruction and (II) the variance of projection.

If we consider the $n$ data points in the matrix $\b{X} \in \mathbb{R}^{d \times n}$, the squared length of reconstruction of the centered data is:
\begin{align*}
||\widehat{\b{X}}||_F^2 &= ||\b{u}\b{u}^\top \breve{\b{X}}||_F^2 = \textbf{tr}\big((\b{u}\b{u}^\top \breve{\b{X}})^\top (\b{u}\b{u}^\top \breve{\b{X}})\big) \\
& = \textbf{tr}(\breve{\b{X}}^\top \b{u} \underbrace{\b{u}^\top \b{u}}_{1} \b{u}^\top \breve{\b{X}}) \overset{(a)}{=} \textbf{tr}(\breve{\b{X}}^\top \b{u} \b{u}^\top \breve{\b{X}}) \\
& \overset{(b)}{=} \textbf{tr}(\b{u}^\top \breve{\b{X}} \breve{\b{X}}^\top \b{u}) \overset{(c)}{=} \b{u}^\top \breve{\b{X}} \breve{\b{X}}^\top \b{u} \overset{(\ref{equation_covariance_matrix})}{=} \b{u}^\top \b{S} \b{u},
\end{align*}
where $\textbf{tr}(.)$ denotes the trace of matrix, $(a)$ is because $\b{u}$ is a unit vector, $(b)$ is because of the cyclic property of the trace, and $(c)$ is because $\b{u}^\top \breve{\b{X}} \breve{\b{X}}^\top \b{u}$ is a scalar.
Hence, we have:
\begin{align}\label{equation_variance_of_projection_oneDirection2}
||\widehat{\b{X}}||_F^2 = \b{u}^\top \b{S} \b{u}.
\end{align}

\subsubsection{Projection Onto a Multi-dimensional Subspace}

In Eq. (\ref{equation_reconstruction_training_SeveralPoints_severalDirections}), if $p>1$, we are projecting the data onto a subspace with dimensionality more than one (spanned by $\{\b{u}_j\}_{j=1}^p$) and then reconstruct back. If the data $\b{X}$ are assumed to be centered, the reconstruction is:
\begin{align*}
\widehat{\b{X}} = \b{U} \b{U}^\top \breve{\b{X}}.
\end{align*}
The squared length (squared Frobenius Norm) of this reconstructed matrix is:
\begin{align*}
||\widehat{\b{X}}||_F^2 &= ||\b{U}\b{U}^\top \breve{\b{X}}||_F^2 = \textbf{tr}\big((\b{U}\b{U}^\top \breve{\b{X}})^\top (\b{U}\b{U}^\top \breve{\b{X}})\big) \\
& = \textbf{tr}(\breve{\b{X}}^\top \b{U} \underbrace{\b{U}^\top \b{U}}_{\b{I}} \b{U}^\top \breve{\b{X}}) \overset{(a)}{=} \textbf{tr}(\breve{\b{X}}^\top \b{U} \b{U}^\top \breve{\b{X}}) \\
& \overset{(b)}{=} \textbf{tr}(\b{U}^\top \breve{\b{X}} \breve{\b{X}}^\top \b{U}) \overset{(\ref{equation_covariance_matrix})}{=} \textbf{tr}(\b{U}^\top \b{S}\, \b{U}),
\end{align*}
where $(a)$ is because $\b{U}$ is an orthogonal matrix (its columns are orthonormal) and $(b)$ is because of the cyclic property of trace.
Thus, we have:
\begin{align}\label{equation_variance_of_projection_SeveralDirections}
||\widehat{\b{X}}||_F^2 = \textbf{tr}(\b{U}^\top \b{S}\, \b{U}).
\end{align}

\section{Fisher Discriminant Analysis}

\subsection{One-dimensional Subspace}

\subsubsection{Scatters in Two-Class Case}

Assume we have two classes, $\{\b{x}_i^{(1)}\}_{i=1}^{n_1}$ and $\{\b{x}_i^{(2)}\}_{i=1}^{n_2}$, where $n_1$ and $n_2$ denote the sample size of the first and second class, respectively, and $\b{x}_i^{(j)}$ denotes the $i$-th instance of the $j$-th class. 

If the data instances of the $j$-th class are projected onto a one-dimensional subspace (vector $\b{u}$) by $\b{u}^\top \b{x}_i^{(j)}$, the mean and the variance of the projected data are $\b{u}^\top \b{\mu}_j$ and $\b{u}^\top \b{S}_j \b{u}$, respectively, where $\b{\mu}_j$ and $\b{S}_j$ are the mean and covariance matrix (scatter) of the $j$-th class. The mean of the $j$-th class is:
\begin{align}\label{equation_mean_of_class}
\mathbb{R}^{d} \ni \b{\mu}_j := \frac{1}{n_j} \sum_{i=1}^{n_j} \b{x}_i^{(j)}.
\end{align}
According to Appendix \ref{section_appendix_metric_learning}, after projection onto the one-dimensional subspace, the distance between the means of classes is:
\begin{align}
\mathbb{R} \ni d_B &:= (\b{u}^\top \b{\mu}_1 - \b{u}^\top \b{\mu}_2)^\top (\b{u}^\top \b{\mu}_1 - \b{u}^\top \b{\mu}_2) \nonumber \\
&= (\b{\mu}_1 - \b{\mu}_2)^\top \b{u}\b{u}^\top (\b{\mu}_1 - \b{\mu}_2) \nonumber \\
&\overset{(a)}{=} \textbf{tr}\big((\b{\mu}_1 - \b{\mu}_2)^\top \b{u}\b{u}^\top (\b{\mu}_1 - \b{\mu}_2)\big) \nonumber \\
&\overset{(b)}{=} \textbf{tr}\big(\b{u}^\top(\b{\mu}_1 - \b{\mu}_2)(\b{\mu}_1 - \b{\mu}_2)^\top \b{u}\big) \nonumber \\
&\overset{(c)}{=} \b{u}^\top (\b{\mu}_1 - \b{\mu}_2)(\b{\mu}_1 - \b{\mu}_2)^\top \b{u} \overset{(d)}{=} \b{u}^\top \b{S}_B\, \b{u}, \label{equation_dB_2classes}
\end{align}
where $(a)$ is because $(\b{\mu}_1 - \b{\mu}_2)^\top \b{u}\b{u}^\top (\b{\mu}_1 - \b{\mu}_2)$ is a scalar, $(b)$ is because of the cyclic property of trace, $(c)$ is because $\b{u}^\top(\b{\mu}_1 - \b{\mu}_2)(\b{\mu}_1 - \b{\mu}_2)^\top \b{u}$ is a scalar, and $(d)$ is because we define:
\begin{align}\label{equation_Fisher_twoClass_between_scatter}
\mathbb{R}^{d \times d} \ni \b{S}_B := (\b{\mu}_1 - \b{\mu}_2)(\b{\mu}_1 - \b{\mu}_2)^\top,
\end{align}
as the \textit{between-scatter} of classes.

The Eq. (\ref{equation_dB_2classes}) can also be interpreted according to Eq. (\ref{equation_variance_of_projection_oneDirection2}): the $d_B$ is the variance of projection of the class means or the squared length of reconstruction of the class means.

We saw that the variance of projection is $\b{u}^\top \b{S}_j \b{u}$ for the $j$-th class. If we add up the variances of projections of the two classes, we have:
\begin{align}
\mathbb{R} \ni d_W &:= \b{u}^\top \b{S}_1 \b{u} + \b{u}^\top \b{S}_2 \b{u} = \b{u}^\top (\b{S}_1 + \b{S}_2)\, \b{u} \nonumber \\
&\overset{(a)}{=} \b{u}^\top \b{S}_W\, \b{u},
\end{align}
where:
\begin{align}\label{equation_within_scatter_twoClasses}
\mathbb{R}^{d \times d} \ni \b{S}_W := \b{S}_1 + \b{S}_2,
\end{align}
is the \textit{within-scatter} of classes.
According to Eq. (\ref{equation_variance_of_projection_oneDirection2}), the $d_W$ is the summation of projection variance of class instances or the summation of the reconstruction length of class instances.

\subsubsection{Scatters in Multi-Class Case: Variant 1}

Assume $\{\b{x}_i^{(j)}\}_{i=1}^{n_j}$ are the instances of the $j$-th class where we have multiple classes.
In this case, the \textit{between-scatter} is defined as:
\begin{align}\label{equation_between_scatter_multipleClasses}
\mathbb{R}^{d \times d} \ni \b{S}_B := \sum_{j=1}^c (\b{\mu}_j - \b{\mu}) (\b{\mu}_j - \b{\mu})^\top,
\end{align}
where $c$ is the number of classes and:
\begin{align}\label{equation_total_mean}
\mathbb{R}^{d} \ni \b{\mu} := \frac{1}{\sum_{k=1}^c n_k} \sum_{j=1}^c n_j\, \b{\mu}_j = \frac{1}{n} \sum_{i=1}^n \b{x}_i,
\end{align}
is the weighted mean of means of classes or the total mean of data.

It is noteworthy that some researches define the between-scatter in a weighted way:
\begin{align}\label{equation_between_scatter_multipleClasses_2}
\mathbb{R}^{d \times d} \ni \b{S}_B := \sum_{j=1}^c n_j (\b{\mu}_j - \b{\mu}) (\b{\mu}_j - \b{\mu})^\top.
\end{align}

If we extend the Eq. (\ref{equation_within_scatter_twoClasses}) to $c$ number of classes, the \textit{within-scatter} is defined as:
\begin{align}
\mathbb{R}^{d \times d} \ni \b{S}_W &:= \sum_{j=1}^c \b{S}_j \\
&\overset{(\ref{equation_covariance_matrix})}{=} \sum_{j=1}^c \sum_{i=1}^{n_j} (\b{x}_i^{(j)} - \b{\mu}_j) (\b{x}_i^{(j)} - \b{\mu}_j)^\top, \label{equation_within_scatter_multipleClasses}
\end{align}
where $n_j$ is the sample size of the $j$-th class.

In this case, the $d_B$ and $d_W$ are:
\begin{align}
&\mathbb{R} \ni d_B := \b{u}^\top \b{S}_B\, \b{u}, \\
&\mathbb{R} \ni d_W := \b{u}^\top \b{S}_W\, \b{u},
\end{align}
where $\b{S}_B$ and $\b{S}_W$ are Eqs. (\ref{equation_between_scatter_multipleClasses}) and (\ref{equation_within_scatter_multipleClasses}).

\subsubsection{Scatters in Multi-Class Case: Variant 2}

There is another variant for multi-class case in FDA. In this variant, the within-scatter is the same as Eq. (\ref{equation_within_scatter_multipleClasses}). 
The between-scatter is, however, different.

The \textit{total-scatter} is defined as the covariance matrix of the whole data, regardless of classes \cite{welling2005fisher}:
\begin{align}\label{equation_fisher_total_scatter}
\mathbb{R}^{d \times d} \ni \b{S}_T := \frac{1}{n} \sum_{i=1}^n (\b{x}_i - \b{\mu}) (\b{x}_i - \b{\mu})^\top,
\end{align}
where the total mean $\b{\mu}$ is the Eq. (\ref{equation_total_mean}). We can also use the scaled total-scatter by dropping the $1/n$ factor.
On the other hand, the total scatter is equal to the summation of the within- and between-scatters:
\begin{align}
\b{S}_T = \b{S}_W + \b{S}_B.
\end{align}
Therefore, the between-scatter, in this variant, is obtained as:
\begin{align}\label{equation_S_B_and_S_T}
\b{S}_B := \b{S}_T - \b{S}_W.
\end{align}

\subsubsection{Fisher Subspace: Variant 1}

In FDA, we want to maximize the projection variance (scatter) of means of classes and minimize the projection variance (scatter) of class instances. In other words, we want to maximize $d_B$ and minimize $d_W$. 
The reason is that after projection, we want the within scatter of every class to be small and the between scatter of classes to be large; therefore, the instances of every class get close to one another and the classes get far from each other. 
The two mentioned optimization problems are:
\begin{align}
&\underset{\b{u}}{\text{maximize}} ~~~ d_B(\b{u}), \label{equation_optimization_d_B} \\
&\underset{\b{u}}{\text{minimize}} ~~~ d_W(\b{u}). \label{equation_optimization_d_W}
\end{align}
We can merge these two optimization problems as a regularized optimization problem:
\begin{align}
&\underset{\b{u}}{\text{maximize}} ~~~ d_B(\b{u}) - \alpha\, d_W(\b{u}), 
\end{align}
where $\alpha>0$ is the regularization parameter. 
Another way of merging Eqs. (\ref{equation_optimization_d_B}) and (\ref{equation_optimization_d_W}) is:
\begin{align}\label{equation_Fisher_optimization_oneDirection_1}
&\underset{\b{u}}{\text{maximize}} ~~~ f(\b{u}) := \frac{d_B(\b{u})}{d_W(\b{u})} = \frac{\b{u}^\top \b{S}_B\, \b{u}}{\b{u}^\top \b{S}_W\, \b{u}}, 
\end{align}
where $f(\b{u}) \in \mathbb{R}$ is referred to as the \textit{Fisher criterion} \cite{xu2006analysis}.
The Fisher criterion is a generalized Rayleigh-Ritz quotient (see Appendix \ref{section_appendix_rayleigh_ritz_quotient}):
\begin{align}
f(\b{u}) \overset{(\ref{equation_generalized_rayleigh_ritz_quotient})}{=} R(\b{S}_B, \b{S}_W; \b{u}).
\end{align}
According to Eq. (\ref{equation_generalized_rayleigh_ritz_quotient_optimization}) in Appendix \ref{section_appendix_rayleigh_ritz_quotient}, the optimization in Eq. (\ref{equation_Fisher_optimization_oneDirection_1}) is equivalent to:
\begin{equation}\label{equation_Fisher_optimization_oneDirection_2}
\begin{aligned}
& \underset{\b{u}}{\text{maximize}}
& & \b{u}^\top \b{S}_B\, \b{u} \\
& \text{subject to}
& & \b{u}^\top \b{S}_W\, \b{u} = 1.
\end{aligned}
\end{equation}
The Lagrangian \cite{boyd2004convex} is: 
\begin{align*}
\mathcal{L} = \b{w}^\top \b{S}_B\, \b{w} - \lambda (\b{w}^\top \b{S}_W\, \b{w} - 1),
\end{align*}
where $\lambda$ is the Lagrange multiplier. Equating the derivative of $\mathcal{L}$ to zero gives: 
\begin{align}
&\mathbb{R}^d \ni \frac{\partial \mathcal{L}}{\partial \b{u}} = 2\,\b{S}_B\,\b{u} - 2\,\lambda\, \b{S}_W\, \b{u} \overset{\text{set}}{=} \b{0} \nonumber \\ 
&  \implies 2\,\b{S}_B\,\b{u} = 2\,\lambda\, \b{S}_W\, \b{u} \implies \b{S}_B\,\b{u} = \lambda\, \b{S}_W\, \b{u},
\end{align}
which is a generalized eigenvalue problem $(\b{S}_B, \b{S}_W)$ according to \cite{ghojogh2019eigenvalue}. The $\b{u}$ is the eigenvector with the largest eigenvalue (because the optimization is maximization) and the $\lambda$ is the corresponding eigenvalue.
The $\b{u}$ is referred to as the \textit{Fisher direction} or \textit{Fisher axis}.
The projection and reconstruction are according to Eqs. (\ref{equation_projection_training_SeveralPoints_severalDirections}) and (\ref{equation_reconstruction_training_SeveralPoints_severalDirections}), respectively, where $\b{u} \in \mathbb{R}^d$ is used instead of $\b{U} \in \mathbb{R}^{d \times p}$.
The out-of-sample projection and reconstruction are according to Eqs. (\ref{equation_projection_outOfSample_SeveralPoints_severalDirections}) and (\ref{equation_reconstruction_outOfSample_SeveralPoints_severalDirections}), respectively, with $\b{u}$ rather than $\b{U}$.

One possible solution to the generalized eigenvalue problem $(\b{S}_B, \b{S}_W)$ is \cite{ghojogh2019eigenvalue}:
\begin{align}
&\b{S}_B\,\b{u} = \lambda\, \b{S}_W\, \b{u} \implies \b{S}_W^{-1} \b{S}_B\,\b{u} = \lambda\, \b{u} \nonumber \\
&\implies
\b{u} = \textbf{eig}(\b{S}_W^{-1} \b{S}_B), \label{equation_Fisher_solution_oneDirection}
\end{align}
where $\textbf{eig}(.)$ denotes the eigenvector of the matrix with the largest eigenvalue.
Although the solution in Eq. (\ref{equation_Fisher_solution_oneDirection}) is a little dirty \cite{ghojogh2019eigenvalue} because $\b{S}_w$ might be singular and not invertible, but this solution is very common for FDA.
In some researches, the diagonal of $\b{S}_W$ is strengthened slightly to make it full rank and invertible \cite{ghojogh2019eigenvalue}:
\begin{align}
\b{u} = \textbf{eig}((\b{S}_W + \varepsilon \b{I})^{-1} \b{S}_B),
\end{align}
where $\varepsilon$ is a very small positive number, large enough to make $\b{S}_W$ full rank.

In a future section, we will cover robust FDA which tackles this problem. On the other hand, the generalized eigenvalue problem $(\b{S}_B, \b{S}_W)$ has a rigorous solution \cite{ghojogh2019eigenvalue,web_generalized_eigenproblem} which does not require non-singularity of $\b{S}_W$. 

Another way to solve the optimization in Eq. (\ref{equation_Fisher_optimization_oneDirection_1}) is taking derivative from the Fisher criterion:
\begin{align}
&\mathbb{R}^d \ni \frac{\partial f(\b{u})}{\partial \b{u}} = \frac{1}{(\b{u}^\top \b{S}_W\, \b{u})^2} \times \nonumber \\
&~~~~~~~ \Big[ (\b{u}^\top \b{S}_W\, \b{u})(2\b{S}_B \b{u}) - (\b{u}^\top \b{S}_B\, \b{u})(2\b{S}_W \b{u}) \Big] \overset{\text{set}}{=} \b{0} \nonumber \\
&\overset{(a)}{\implies} \b{S}_B\, \b{u} = \frac{\b{u}^\top \b{S}_B\, \b{u}}{\b{u}^\top \b{S}_W\, \b{u}}\, \b{S}_W\, \b{u}, \label{equation_Fisher_solution_oneDirection_2}
\end{align}
where $(a)$ is because $\b{u}^\top \b{S}_W\, \b{u}$ is a scalar.
The Eq. (\ref{equation_Fisher_solution_oneDirection_2}) which is a generalized eigenvalue problem $(\b{S}_B, \b{S}_W)$ \cite{ghojogh2019eigenvalue} with $\b{u}$ and $(\b{u}^\top \b{S}_B\, \b{u}) / (\b{u}^\top \b{S}_W\, \b{u})$ as the eigenvector with the largest eigenvalue (because the optimization is maximization) and the corresponding eigenvalue, respectively. Therefore, \textit{the Fisher criterion is the eigenvalue of the Fisher direction.} 

\subsubsection{Fisher Subspace: Variant 2}

Another way to find the FDA direction is to 
consider another version of Fisher criterion. According to Eq. (\ref{equation_S_B_and_S_T}) for $\b{S}_B$, the Fisher criterion becomes \cite{welling2005fisher}:
\begin{align}
f(\b{u}) &= \frac{\b{u}^\top \b{S}_B\, \b{u}}{\b{u}^\top \b{S}_W\, \b{u}} \overset{(\ref{equation_S_B_and_S_T})}{=} \frac{\b{u}^\top (\b{S}_T - \b{S}_W)\, \b{u}}{\b{u}^\top \b{S}_W\, \b{u}} \nonumber \\
&=  \frac{\b{u}^\top \b{S}_T\, \b{u} - \b{u}^\top \b{S}_W\, \b{u}}{\b{u}^\top \b{S}_W\, \b{u}} = \frac{\b{u}^\top \b{S}_T\, \b{u}}{\b{u}^\top \b{S}_W\, \b{u}} - 1. \label{equation_Fisher_criterion_variant2_oneDimensional}
\end{align}
The $-1$ is a constant and is dropped in the optimization; therefore:
\begin{equation}\label{equation_optimization_FDA_with_S_T_oneDirection}
\begin{aligned}
& \underset{\b{u}}{\text{maximize}}
& & \b{u}^\top \b{S}_T\, \b{u} \\
& \text{subject to}
& & \b{u}^\top \b{S}_W\, \b{u} = 1,
\end{aligned}
\end{equation}
whose solution is similarly obtained as:
\begin{align}
\b{S}_T\,\b{u} = \lambda\, \b{S}_W\, \b{u},
\end{align}
which is a generalized eigenvalue problem $(\b{S}_T, \b{S}_W)$ according to \cite{ghojogh2019eigenvalue}.

\subsection{Multi-dimensional Subspace}

In case the Fisher subspace is the span of several Fisher directions, $\{\b{u}_j\}_{j=1}^p$ where $\b{u}_j \in \mathbb{R}^d$, the $d_B$ and $d_W$ are defined as:
\begin{align}
&\mathbb{R} \ni d_B := \textbf{tr}(\b{U}^\top \b{S}_B\, \b{U}), \\
&\mathbb{R} \ni d_W := \textbf{tr}(\b{U}^\top \b{S}_W\, \b{U}),
\end{align}
where $\mathbb{R}^{d \times p} \ni \b{U} = [\b{u}_1, \dots, \b{u}_p]$.
In this case, maximizing the \textit{Fisher criterion} is:
\begin{align}\label{equation_Fisher_optimization_SeveralDirections_1}
&\underset{\b{U}}{\text{maximize}} ~~~ f(\b{U}) := \frac{d_B(\b{U})}{d_W(\b{U})} = \frac{\textbf{tr}(\b{U}^\top \b{S}_B\, \b{U})}{\textbf{tr}(\b{U}^\top \b{S}_W\, \b{U})}. 
\end{align}
The Fisher criterion $f(\b{U})$ is a generalized Rayleigh-Ritz quotient (see Appendix \ref{section_appendix_rayleigh_ritz_quotient}).
According to Eq. (\ref{equation_generalized_rayleigh_ritz_quotient_optimization}) in Appendix \ref{section_appendix_rayleigh_ritz_quotient}, the optimization in Eq. (\ref{equation_Fisher_optimization_SeveralDirections_1}) is \textit{approximately} equivalent to:
\begin{equation}\label{equation_Fisher_optimization_severalDirections_2}
\begin{aligned}
& \underset{\b{U}}{\text{maximize}}
& & \textbf{tr}(\b{U}^\top \b{S}_B\, \b{U}) \\
& \text{subject to}
& & \b{U}^\top \b{S}_W\, \b{U} = \b{I}.
\end{aligned}
\end{equation}
Note that Eq. (\ref{equation_generalized_rayleigh_ritz_quotient_optimization}) is exactly true for one projection vector $\b{u}$ but it approximately holds for the projection matrix $\b{U}$ having multiple projection directions. 
The Lagrangian \cite{boyd2004convex} is: 
\begin{align*}
\mathcal{L} = \textbf{tr}(\b{U}^\top \b{S}_B\, \b{U}) - \textbf{tr}\big(\b{\Lambda}^\top (\b{U}^\top \b{S}_W\, \b{U} - \b{I})\big),
\end{align*}
where $\b{\Lambda} \in \mathbb{R}^{d \times d}$ is a diagonal matrix whose diagonal entries are the Lagrange multipliers. Equating the derivative of $\mathcal{L}$ to zero gives: 
\begin{align}
&\mathbb{R}^{d \times p} \ni \frac{\partial \mathcal{L}}{\partial \b{U}} = 2\,\b{S}_B\,\b{U} - 2\,\b{S}_W\, \b{U} \b{\Lambda} \overset{\text{set}}{=} \b{0} \nonumber \\ 
&  \implies 2\,\b{S}_B\,\b{U} = 2\, \b{S}_W\, \b{U} \b{\Lambda} \implies \b{S}_B\,\b{U} = \b{S}_W\, \b{U} \b{\Lambda},
\end{align}
which is a generalized eigenvalue problem $(\b{S}_B, \b{S}_W)$ according to \cite{ghojogh2019eigenvalue}. The columns of $\b{U}$ are the eigenvectors sorted by largest to smallest eigenvalues (because the optimization is maximization) and the diagonal entries of $\b{\Lambda}$ are the corresponding eigenvalues.
The columns of $\b{U}$ are referred to as the \textit{Fisher directions} or \textit{Fisher axes}.
The projection and reconstruction are according to Eqs. (\ref{equation_projection_training_SeveralPoints_severalDirections}) and (\ref{equation_reconstruction_training_SeveralPoints_severalDirections}), respectively.
The out-of-sample projection and reconstruction are according to Eqs. (\ref{equation_projection_outOfSample_SeveralPoints_severalDirections}) and (\ref{equation_reconstruction_outOfSample_SeveralPoints_severalDirections}), respectively.

One possible solution to the generalized eigenvalue problem $(\b{S}_B, \b{S}_W)$ is \cite{ghojogh2019eigenvalue}:
\begin{align}
&\b{S}_B\,\b{U} = \b{S}_W\, \b{U} \b{\Lambda} \implies \b{S}_W^{-1} \b{S}_B\,\b{U} = \b{U} \b{\Lambda} \nonumber \\
&\implies
\b{U} = \textbf{eig}(\b{S}_W^{-1} \b{S}_B), \label{equation_Fisher_solution_severalDirections}
\end{align}
where $\textbf{eig}(.)$ denotes the eigenvectors of the matrix stacked column-wise.
Again, we can have \cite{ghojogh2019eigenvalue}:
\begin{align}
\b{U} = \textbf{eig}((\b{S}_W + \varepsilon \b{I})^{-1} \b{S}_B),
\end{align}

Another way to solve the optimization in Eq. (\ref{equation_Fisher_optimization_SeveralDirections_1}) is taking derivative from the Fisher criterion:
\begin{align}
&\mathbb{R}^{d \times p} \ni \frac{\partial f(\b{U})}{\partial \b{U}} = \frac{1}{\big( \textbf{tr}(\b{U}^\top \b{S}_W\, \b{U})\big)^2} \times \nonumber \\
& \Big[  \textbf{tr}(\b{U}^\top \b{S}_W\, \b{U})(2\b{S}_B \b{U}) -  \textbf{tr}(\b{U}^\top \b{S}_B\, \b{U})(2\b{S}_W \b{U}) \Big] \overset{\text{set}}{=} \b{0} \nonumber \\
&\overset{(a)}{\implies} \b{S}_B\, \b{U} = \frac{\textbf{tr}(\b{U}^\top \b{S}_B\, \b{U})}{\textbf{tr}(\b{U}^\top \b{S}_W\, \b{U})}\, \b{S}_W\, \b{U}, \label{equation_Fisher_solution_SeveralDirections_2}
\end{align}
where $(a)$ is because $\textbf{tr}(\b{U}^\top \b{S}_W\, \b{U})$ is a scalar.
The Eq. (\ref{equation_Fisher_solution_SeveralDirections_2}) which is a generalized eigenvalue problem $(\b{S}_B, \b{S}_W)$ \cite{ghojogh2019eigenvalue} with columns of $\b{U}$ as the eigenvectors and $(\b{u}_j^\top \b{S}_B\, \b{u}_j) / (\b{u}_j^\top \b{S}_W\, \b{u}_j)$ as the $j$-th largest eigenvalue (because the optimization is maximization).

Again, another way to find the FDA directions is to 
consider another version of Fisher criterion. According to Eq. (\ref{equation_S_B_and_S_T}) for $\b{S}_B$, the Fisher criterion becomes \cite{welling2005fisher}:
\begin{align}\label{equation_Fisher_criterion_variant2_multiDimensional}
f(\b{U}) &= \frac{\textbf{tr}\big(\b{U}^\top (\b{S}_T - \b{S}_W)\, \b{U}\big)}{\textbf{tr}(\b{U}^\top \b{S}_W\, \b{U})} = \frac{\textbf{tr}(\b{U}^\top \b{S}_T\, \b{U})}{\textbf{tr}(\b{U}^\top \b{S}_W\, \b{U})} - 1.
\end{align}
The $-1$ is a constant and is dropped in the optimization; therefore:
\begin{equation}\label{equation_optimization_FDA_with_S_T_multiDimensional}
\begin{aligned}
& \underset{\b{U}}{\text{maximize}}
& & \textbf{tr}(\b{U}^\top \b{S}_T\, \b{U}) \\
& \text{subject to}
& & \b{U}^\top \b{S}_W\, \b{U} = \b{I},
\end{aligned}
\end{equation}
whose solution is similarly obtained as:
\begin{align}
\b{S}_T\,\b{U} = \b{S}_W\, \b{U} \b{\Lambda},
\end{align}
which is a generalized eigenvalue problem $(\b{S}_T, \b{S}_W)$ according to \cite{ghojogh2019eigenvalue}.

\subsection{Discussion on Dimensionality of the Fisher Subspace}\label{section_Fisher_dimensionality}

In general, the rank of a covariance (scatter) matrix over the $d$-dimensional data with sample size $n$ is at most $\min(d, n-1)$. The $d$ is because the covariance matrix is a $d \times d$ matrix and the $n$ is because we iterate over $n$ data instances for calculating the covariance matrix. The $-1$ is because of subtracting the mean in calculation of the covariance matrix. For clarification, assume we only have one instance which becomes zero after removing the mean. This makes the covariance matrix a zero matrix. 

According to Eq. (\ref{equation_within_scatter_multipleClasses}), the rank of the $\b{S}_W$ is at most $\min(d, n-1)$ because all the instances of all the classes are considered. Hence, the rank of $\b{S}_W$ is also at most $\min(d, n-1)$.
According to Eq. (\ref{equation_between_scatter_multipleClasses}), the rank of the $\b{S}_B$ is at most $\min(d, c-1)$ because we have $c$ iterations in its calculation. 

In Eq. (\ref{equation_Fisher_solution_severalDirections}), we have $\b{S}_W^{-1} \b{S}_B$ whose rank is:
\begin{align}
&\textbf{rank}(\b{S}_W^{-1} \b{S}_B) \leq \min\big(\textbf{rank}(\b{S}_W^{-1}), \textbf{rank}(\b{S}_B)\big) \nonumber \\
&\leq \min\big(\min(d, n-1), \min(d, c-1)\big) \nonumber \\
&= \min(d, n-1, c-1) \overset{(a)}{=} c-1,
\end{align}
where $(a)$ is because we usually have $c < d,n$.
Therefore, the rank of $\b{S}_W^{-1} \b{S}_B$ is limited because of the rank of $\b{S}_B$ which is at most $c-1$. 

According to Eq. (\ref{equation_Fisher_solution_severalDirections}), the $c-1$ leading eigenvalues will be valid and the rest are zero or very small. Therefore, the $p$, which is the dimensionality of the Fisher subspace, is at most $c-1$. The $c-1$ leading eigenvectors are considered as the Fisher directions and the rest of eigenvectors are invalid and ignored. 

\section{Interpretation of FDA: The Example of a Man with Weak Eyes}

In this section, we interpret the FDA using a real-life example in order to better understand the essence of Fisher's method. 
Consider a man which has two eye problems: (1) he is color-blind and (2) his eyes are also very weak. 

Suppose there are two sets of balls with red and blue colors. The man wants to discriminate the balls into red and blue classes; however, he needs help because of his eye problems. 

First, consider his color-blindness. In order to help him, we separate the balls into two sets of red and blue. In other words, we increase the distances of the balls with different colors to give him a clue that which balls belong to the same class. This means that we are increasing the between-scatter of the two classes to help him.

Second, consider his very weak eyes. although the balls with different colors are almost separated, everything is blue to him. Thus, we put the balls of the same color closer to one another. In other words, we decrease the within-scatter of every class. In this way, the man sees every class as almost one blurry ball so he can discriminate the classes better. 

Recall Eq. (\ref{equation_Fisher_solution_severalDirections}) which includes $\b{S}_W^{-1} \b{S}_B$. The $\b{S}_B$ implies that we want to increase the between-scatter as we did in the first help. The $\b{S}_W^{-1}$ implies that we want to decrease the within-scatter as done in the second help to the man.
In conclusion, FDA increases the between-scatter and decreases the within-scatter (collapses each class \cite{globerson2006metric}), at the same time, for better discrimination of the classes. 

\section{Robust Fisher Discriminant Analysis}

Robust FDA (RFDA) \cite{deng2007robust,guo2015feature}, has also addressed the problem of singularity (or close to singularity) of $\b{S}_W$. In RFDA, the $\b{S}_W$ is decomposed using eigenvalue decomposition \cite{ghojogh2019eigenvalue}:
\begin{align}
\b{S}_W = \b{\Phi}^\top \b{\Lambda} \b{\Phi},
\end{align}
where $\b{\Phi}$ and $\b{\Lambda} = \textbf{diag}([\lambda_1, \dots, \lambda_d]^\top)$ include the eigenvectors and eigenvalues of $\b{S}_W$, respectively. 
The eigenvalues are sorted as $\lambda_1 \geq \dots \geq \lambda_d$ and the eigenvectors (columns of $\b{\Phi}$) are sorted accordingly. 
If $\b{S}_W$ is close to singularity, the first $d'$ eigenvalues are valid and the rest $(d-d')$ eigenvalues are either very small or zero. 
The appropriate $d'$ is obtained as:
\begin{align}
d' := \arg\min_{m} \bigg(\frac{\sum_{j=1}^m \lambda_j}{\sum_{k=1}^d \lambda_k} \geq 0.98 \bigg).
\end{align}
In RFDA, the $(d - d')$ invalid eigenvalues are replaced with $\lambda_*$:
\begin{align}
\mathbb{R}^{d \times d} \ni \b{\Lambda}' := \textbf{diag}([\lambda_1, \dots, \lambda_{d'}, \lambda_*, \dots, \lambda_*]^\top),
\end{align}
where \cite{deng2007robust}:
\begin{align}
\lambda_* := \frac{1}{d - d'} \sum_{j=d'+1}^d \lambda_j.
\end{align}
Hence, the $\b{S}_W$ is replaced with $\b{S}'_W$:
\begin{align}
\mathbb{R}^{d \times d} \ni \b{S}'_W := \b{\Phi}^\top \b{\Lambda}' \b{\Phi},
\end{align}
and the robust Fisher directions are the eigenvectors of the generalized eigenvalue problem $(\b{S}_B, \b{S}'_W)$ \cite{ghojogh2019eigenvalue}.

\begin{figure}[!t]
\centering
\includegraphics[width=2.6in]{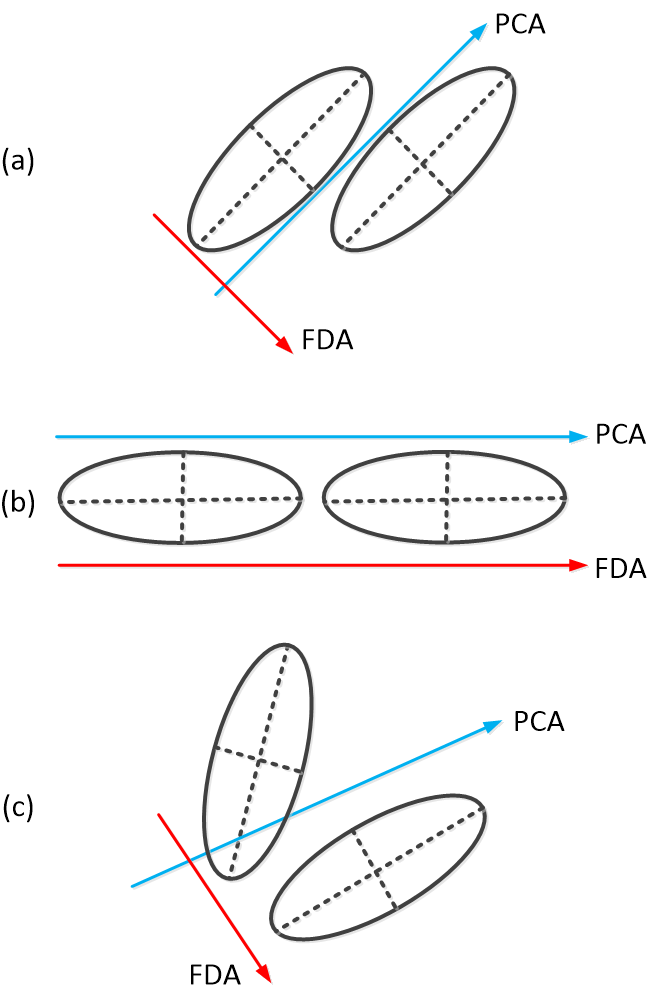}
\caption{Comparison of FDA and PCA directions for two-dimensional data with two classes: (a) a case where FDA and PCA are orthogonal, (b) a case where FDA and PCA are equivalent (parallel), and (c) a case between the two extreme cases of (a) and (b).}
\label{figure_PCA_FDA_comparison}
\end{figure}

\section{Comparison of FDA and PCA Directions}

The FDA directions capture the directions where the instances of different classes fall apart and the instances in one class fall close to each other. On the other hand, the PCA directions capture the directions where the data have maximum variance (spread) regardless of the classes \cite{ghojogh2019unsupervised}. In some datasets, the FDA and PCA are orthogonal and in some datasets, they are parallel. Other cases between these two extreme cases can happen for some datasets. This depends on the spread of classes in the dataset. Figure \ref{figure_PCA_FDA_comparison} shows these cases for some two-dimensional datasets.

Moreover, considering the Eq. (\ref{equation_S_B_and_S_T}) for $\b{S}_B$, the Fisher criterion becomes Eqs. (\ref{equation_Fisher_criterion_variant2_oneDimensional}) and (\ref{equation_Fisher_criterion_variant2_multiDimensional}) for one-dimensional and multi-dimensional Fisher subspaces, respectively. 
In these equations, the $-1$ is a constant and is dropped in the optimization. This has an important message about FDA: \textit{the Fisher direction is maximizing the total variance (spread) of data, as also done in PCA, while at the same time, it minimizes the within-scatters of classes (by making use of the class labels).}
In other words, the optimization of FDA is equivalent to (we repeat Eq. (\ref{equation_optimization_FDA_with_S_T_multiDimensional}) here):
\begin{equation}\label{equation_optimization_FDA_with_S_T_multiDimensional_2}
\begin{aligned}
& \underset{\b{U}}{\text{maximize}}
& & \textbf{tr}(\b{U}^\top \b{S}_T\, \b{U}) \\
& \text{subject to}
& & \b{U}^\top \b{S}_W\, \b{U} = \b{I},
\end{aligned}
\end{equation}
while the optimization of the PCA is \cite{ghojogh2019unsupervised}:
\begin{equation}\label{equation_optimization_PCA_with_S_T}
\begin{aligned}
& \underset{\b{U}}{\text{maximize}}
& & \textbf{tr}(\b{U}^\top \b{S}_T\, \b{U}) \\
& \text{subject to}
& & \b{U}^\top \b{U} = \b{I}.
\end{aligned}
\end{equation}
The solutions to Eqs. (\ref{equation_optimization_FDA_with_S_T_multiDimensional_2}) and (\ref{equation_optimization_PCA_with_S_T}) are the generalized eigenvalue problem $(\b{S}_T, \b{S}_W)$ and the eigenvalue problem for $\b{S}_T$, respectively \cite{ghojogh2019eigenvalue}.

\section{FDA $\overset{?}{\equiv}$ LDA}

The FDA is also referred to as Linear Discriminant Analysis (LDA) and Fisher LDA (FLDA). 
Note that FDA is a manifold (subspace) learning method and LDA \cite{ghojogh2019linear} is a classification method. However, LDA can be seen as a metric learning method \cite{ghojogh2019linear} and as metric learning is a subspace learning method (see Appendix \ref{section_appendix_metric_learning}), there is a connection between FDA and LDA.

We know that FDA is a projection-based subspace learning method. Consider the projection vector $\b{u}$. According to Eq. (\ref{equation_projection_training_onePoint_severalDirections}), the projection of data $\b{x}$ is:
\begin{align}\label{equation_FDA_projection}
\b{x} \mapsto \b{u}^\top \b{x},
\end{align}
which can be done for all the data instances of every class. Thus, the mean and the covariance matrix of the class are transformed as:
\begin{align}
& \b{\mu} \mapsto \b{u}^\top \b{\mu}, \\
& \b{\Sigma} \mapsto \b{u}^\top \b{\Sigma}\, \b{u},
\end{align}
respectively, because of characteristics of mean and variance. 

According to Eq. (\ref{equation_Fisher_optimization_oneDirection_1}), the Fisher criterion is the ratio of the between-class variance, $\sigma^2_b$, and within-class variance, $\sigma^2_w$:
\begin{align}
f := \frac{\sigma^2_b}{\sigma^2_w} = \frac{(\b{u}^\top \b{\mu}_2 - \b{u}^\top \b{\mu}_1)^2}{\b{u}^\top \b{\Sigma}_2\, \b{u} + \b{u}^\top \b{\Sigma}_1\, \b{u}} = \frac{\big(\b{u}^\top (\b{\mu}_2 - \b{\mu}_1)\big)^2}{\b{u}^\top (\b{\Sigma}_2 + \b{\Sigma}_1)\, \b{u}},
\end{align}
where $\b{\mu}_1$ and $\b{\mu}_2$ are the means of the two classes and $\b{\Sigma}_1$ and $\b{\Sigma}_2$ are the covariances of the two classes. 
The FDA maximizes the Fisher criterion:
\begin{equation}\label{equation_Fisher_optimization}
\begin{aligned}
& \underset{\b{u}}{\text{maximize}}
& & \frac{\big(\b{u}^\top (\b{\mu}_2 - \b{\mu}_1)\big)^2}{\b{u}^\top (\b{\Sigma}_2 + \b{\Sigma}_1)\, \b{u}}, \\
\end{aligned}
\end{equation}
which can be restated as:
\begin{equation}\label{equation_Fisher_optimization_2}
\begin{aligned}
& \underset{\b{u}}{\text{maximize}}
& & \big(\b{u}^\top (\b{\mu}_2 - \b{\mu}_1)\big)^2, \\
& \text{subject to}
& & \b{u}^\top (\b{\Sigma}_2 + \b{\Sigma}_1)\, \b{u} = 1,
\end{aligned}
\end{equation}
according to Rayleigh-Ritz quotient method \cite{croot2005rayleigh}.
The Lagrangian \cite{boyd2004convex} is: 
\begin{align*}
\mathcal{L} = \big(\b{u}^\top (\b{\mu}_2 - \b{\mu}_1)\big)^2 - \lambda \big(\b{u}^\top (\b{\Sigma}_2 + \b{\Sigma}_1)\, \b{u} - 1\big),
\end{align*}
where $\lambda$ is the Lagrange multiplier. Equating the derivative of $\mathcal{L}$ to zero gives: 
\begin{align*}
&\frac{\partial \mathcal{L}}{\partial \b{u}} = 2\,(\b{\mu}_2 - \b{\mu}_1) (\b{\mu}_2 - \b{\mu}_1)^\top \b{u} - 2\,\lambda\, (\b{\Sigma}_2 + \b{\Sigma}_1)\, \b{u} \overset{\text{set}}{=} \b{0} \\ 
& \implies (\b{\mu}_2 - \b{\mu}_1) (\b{\mu}_2 - \b{\mu}_1)^\top \b{u} = \lambda\, (\b{\Sigma}_2 + \b{\Sigma}_1)\, \b{u},
\end{align*}
which is a generalized eigenvalue problem $\big((\b{\mu}_2 - \b{\mu}_1) (\b{\mu}_2 - \b{\mu}_1)^\top, (\b{\Sigma}_2 + \b{\Sigma}_1)\big)$ according to \cite{ghojogh2019eigenvalue}. The projection vector is the eigenvector of $(\b{\Sigma}_2 + \b{\Sigma}_1)^{-1} (\b{\mu}_2 - \b{\mu}_1) (\b{\mu}_2 - \b{\mu}_1)^\top$; therefore, we can say:
\begin{align}\label{equation_u_LDA_and_FDA}
\b{u} \propto (\b{\Sigma}_2 + \b{\Sigma}_1)^{-1} (\b{\mu}_2 - \b{\mu}_1) (\b{\mu}_2 - \b{\mu}_1)^\top.
\end{align}

On the other hand, in LDA, the decision function is \cite{ghojogh2019linear}:
\begin{equation}\label{equation_LDA_equal_priors}
\begin{aligned}
&2\,\big(\b{\Sigma}^{-1} (\b{\mu}_2 - \b{\mu}_1)\big)^\top \b{x} \\
&~~~~~ + \b{\mu}_1^{\top} \b{\Sigma}^{-1} \b{\mu}_1 - \b{\mu}_2^{\top} \b{\Sigma}^{-1} \b{\mu}_2 + 2 \ln(\frac{\pi_1}{\pi_2}) = 0,
\end{aligned}
\end{equation}
where $\pi_1$ and $\pi_2$ are the prior distributions of the two classes. 
Moreover, in LDA, the covariance matrices are assumed to be equal \cite{ghojogh2019linear}: $\b{\Sigma}_1 = \b{\Sigma}_2 = \b{\Sigma}$. Therefore, in LDA, the Eq. (\ref{equation_u_LDA_and_FDA}) becomes \cite{ghojogh2019linear}:
\begin{align}
\b{u} &\propto (2\,\b{\Sigma})^{-1} (\b{\mu}_2 - \b{\mu}_1) (\b{\mu}_2 - \b{\mu}_1)^\top \nonumber \\
&\propto \b{\Sigma}^{-1} (\b{\mu}_2 - \b{\mu}_1) (\b{\mu}_2 - \b{\mu}_1)^\top.
\end{align}
According to Eq. (\ref{equation_FDA_projection}), we have:
\begin{align}\label{equation_FDA_projection_2}
\b{u}^\top \b{x} \propto \big(\b{\Sigma}^{-1} (\b{\mu}_2 - \b{\mu}_1) (\b{\mu}_2 - \b{\mu}_1)^\top \big)^\top \b{x}.
\end{align}
Comparing Eq. (\ref{equation_FDA_projection_2}) with Eq. (\ref{equation_LDA_equal_priors}) shows that LDA and FDA are equivalent up to a scaling factor $\b{\mu}_1^{\top} \b{\Sigma}^{-1} \b{\mu}_1 - \b{\mu}_2^{\top} \b{\Sigma}^{-1} \b{\mu}_2 + 2 \pi_1 / \pi_2$. Note that this term is multiplied as an exponential factor before taking logarithm to obtain Eq. (\ref{equation_LDA_equal_priors}), so this term is a scaling factor (see \cite{ghojogh2019linear} for more details). It should be noted that in manifold (subspace) learning, the scale does not matter because all the distances can scale similarly in the subspace, without impacting the relative distances of points.
Hence, we can say that LDA and FDA are equivalent:
\begin{align}
\text{LDA} \equiv \text{FDA}.
\end{align}
Therefore, \textit{the two subspaces of FDA and LDA are the same subspace.} 
In other words, FDA followed by the use of Euclidean distance for classification in the subspace is equivalent to LDA.
This sheds light on why LDA and FDA are used interchangeably in the literature. 

Note that LDA assumes \textit{one} (and not several) Gaussian for every class \cite{ghojogh2019linear} and so does the FDA because they are equivalent. That is why FDA faces problem for multi-modal data \cite{sugiyama2007dimensionality}.

\section{Fisher Forest}

If the data include several different types of data which may even have different dimensionality. Some examples of these types of data are different key-poses in action human action recognition or different facial expressions that a face can have.
In this case, we can use the \textit{Fisher forest} \cite{ghojogh2017automatic}. Note that forest here does not imply an ensemble of trees but means an ensemble of the Fisher subspaces. 

Let the number of the data types be $z$ and let the dimensionality of the $\ell$-th data type be $d_{|\ell}$. 
We usually have a dataset $\{\b{x}_i\}_{i=1}^n$ where $\b{x}_i \in \mathbb{R}^d$. Every type of data is the whole dataset but having only a subset of the features, i.e., $\{\b{x}_{i|\ell}\}_{i=1}^{n}$ where $\b{x}_{i|\ell} \in \mathbb{R}^{d_{|\ell}}$. The features of the $\ell$-th type are a subset of the features of the dataset, i.e., $d_{|\ell} \leq d$. Note that, we do not necessarily have the same value for $d_{|\ell}$ in all the data types. 
The $i$-th instance of the $j$-th class having the $\ell$-th type is denoted by $\b{x}_{i|\ell}^{(j)}$.

For example, in the key-poses of human action, the important of skeletal joints can be different in various key-poses \cite{ghojogh2017automatic}. Thus, some joints are taken in a specific key-joint and some other are taken in another key-pose. 
Note that a key-pose can have five key-joints but another key-pose can have three key-joints.
Another example is using different landmarks for different facial expressions; for example, eye-brows, lips, and chin for wondering but just lips for smiling. 
As can be seen, Fisher forest can be useful for handling the data types with different features and even dimensionality. 

The between- and within-scatters for the $\ell$-th data types (for all $\ell \in \{1, \dots, z\}$) are defined as \cite{ghojogh2017automatic}:
\begin{align}
&\mathbb{R}^{d_\ell \times d_\ell} \ni \b{S}_{B|\ell} := \sum_{j=1}^c n_j (\b{\mu}_{j|\ell} - \b{\mu}_{|\ell}) (\b{\mu}_{j|\ell} - \b{\mu}_{|\ell})^\top, \\
&\mathbb{R}^{d_\ell \times d_\ell} \ni \b{S}_{W|\ell} := \sum_{j=1}^c \sum_{i=1}^{n_j} (\b{x}_{i|\ell}^{(j)} - \b{\mu}_{j|\ell}) (\b{x}_{i|\ell}^{(j)} - \b{\mu}_{j|\ell})^\top,
\end{align}
where:
\begin{align}
&\mathbb{R}^{d_{|\ell}} \ni \b{\mu}_{j|\ell} := \frac{1}{n_j} \sum_{i=1}^{n_j} \b{x}_{i|\ell}^{(j)}, \\
&\mathbb{R}^{d_{|\ell}} \ni \b{\mu}_{|\ell} := \frac{1}{n} \sum_{i=1}^{n} \b{x}_{i|\ell}.
\end{align}
The $\ell$-th Fisher subspace is spanned by the eigenvectors of $\b{S}_{W|\ell}^{-1} \b{S}_{B|\ell}$.

Hence, $z$ Fisher subspaces are trained.
In the test phase, the data instance is projected onto every subspace. If we want to classify the data instance in the projected subspaces, we will have $z$ classification results after projection onto these $z$ subspaces. We can then use majority voting for a final classification of the data instance \cite{ghojogh2017automatic}. The effectiveness of the majority voting can be explained because of ensemble learning \cite{polikar2012ensemble,ghojogh2019theory}.
We can also normalize the distances in the subspaces of Fisher forest for the sake of classification (see \cite{ghojogh2017automatic} for more details).

\section{Kernel Fisher Discriminant Analysis}

\subsection{Kernels and Hilbert Space}

Suppose that $\b{\phi}: \b{x} \rightarrow \mathcal{H}$ is a function which maps the data $\b{x}$ to Hilbert space (feature space). The $\b{\phi}$ is called \textit{pulling function}. In other words, $\b{x} \mapsto \b{\phi}(\b{x})$. Let $t$ denote the dimensionality of the feature space, i.e., $\b{\phi}(\b{x}) \in \mathbb{R}^t$ while $\b{x} \in \mathbb{R}^d$. Note that we usually have $t \gg d$.

If $\mathcal{X}$ denotes the set of points, i.e., $\b{x} \in \mathcal{X}$, the kernel of two vectors $\b{x}_1$ and $\b{x}_2$ is $k: \mathcal{X} \times \mathcal{X} \rightarrow \mathbb{R}$ and is defined as \cite{hofmann2008kernel,herbrich2001learning}:
\begin{align}\label{equation_kernel_scalar}
k(\b{x}_1, \b{x}_2) := \b{\phi}(\b{x}_1)^\top \b{\phi}(\b{x}_2),
\end{align}
which is a measure of \textit{similarity} between the two vectors because the inner product captures similarity.

We can compute the kernel of two matrices $\b{X}_1 \in \mathbb{R}^{d \times n_1}$ and $\b{X}_2 \in \mathbb{R}^{d \times n_2}$ and have a \textit{kernel matrix} (also called \textit{Gram matrix}):
\begin{align}
\mathbb{R}^{n_1 \times n_2} \ni \b{K}(\b{X}_1, \b{X}_2) := \b{\Phi}(\b{X}_1)^\top \b{\Phi}(\b{X}_2),
\end{align}
where $\b{\Phi}(\b{X}_1) := [\b{\phi}(\b{x}_1), \dots, \b{\phi}(\b{x}_n)] \in \mathbb{R}^{t \times n_1}$ is the matrix of mapped $\b{X}_1$ to the feature space. The $\b{\Phi}(\b{X}_2) \in \mathbb{R}^{t \times n_2}$ is defined similarly. 
We can compute the kernel matrix of dataset $\b{X} \in \mathbb{R}^{d \times n}$ over itself:
\begin{align}\label{equation_kernel_matrix_of_X}
\mathbb{R}^{n \times n} \ni \b{K}_x := \b{K}(\b{X}, \b{X}) = \b{\Phi}(\b{X})^\top \b{\Phi}(\b{X}),
\end{align}
where $\b{\Phi}(\b{X}) := [\b{\phi}(\b{x}_1), \dots, \b{\phi}(\b{x}_n)] \in \mathbb{R}^{t \times n}$ is the pulled (mapped) data.

Note that in kernel methods, the pulled data $\b{\Phi}(\b{X})$ are usually not available and merely the kernel matrix $\b{K}(\b{X}, \b{X})$, which is the inner product of the pulled data with itself, is available.

There exist different types of kernels. Some of the most well-known kernels are:
\begin{align}
&\text{Linear:} ~~ k(\b{x}_1, \b{x}_2) = \b{x}_1^\top \b{x}_2 + c_1, \\
&\text{Polynomial:} ~~ k(\b{x}_1, \b{x}_2) = (c_1\b{x}_1^\top \b{x}_2 + c_2)^{c_3}, \\
&\text{Gaussian:} ~~ k(\b{x}_1, \b{x}_2) = \exp\big(\!-\frac{||\b{x}_1 - \b{x}_2||_2^2}{2\sigma^2}\big), \\
&\text{Sigmoid:} ~~ k(\b{x}_1, \b{x}_2) = \tanh(c_1\b{x}_1^\top\b{x}_2 + c_2), 
\end{align}
where $c_1$, $c_2$, $c_3$, and $\sigma$ are scalar constants. The Gaussian and Sigmoid kernels are also called Radial Basis Function (RBF) and hyperbolic tangent, respectively. Note that the Gaussian kernel can also be written as $\exp\big(\!-\gamma||\b{x}_1 - \b{x}_2||_2^2\big)$ where $\gamma > 0$.

It is noteworthy to mention that in the RBF kernel, the dimensionality of the feature space is infinite. The reason lies in the Maclaurin series expansion (Taylor series expansion at zero) of this kernel:
\begin{align*}
\exp(-\gamma r) \approx 1 - \gamma r + \frac{\gamma^2}{2!} r^2 - \frac{\gamma^3}{3!} r^3 + \dots,
\end{align*}
where $r := ||\b{x}_1 - \b{x}_2||_2^2$, which is infinite dimensional with respect to $r$.

\subsection{One-dimensional Subspace}

\subsubsection{Scatters in Two-Class Case}

The Eq. (\ref{equation_Fisher_twoClass_between_scatter}) in the feature space is:
\begin{align}\label{equation_kernel_Fisher_twoClass_between_scatter}
&\mathbb{R}^{t \times t} \ni \b{\Phi}(\b{S}_B) := \nonumber \\
&~~~~~~~~~~~~~~~ \big(\b{\phi}(\b{\mu}_1) - \b{\phi}(\b{\mu}_2)\big)\big(\b{\phi}(\b{\mu}_1) - \b{\phi}(\b{\mu}_2)\big)^\top,
\end{align}
where the mean of the $j$-th class in the feature space is:
\begin{align}\label{equation_mean_of_class_inFeatureSpace}
\mathbb{R}^{t} \ni \b{\phi}(\b{\mu}_j) := \frac{1}{n_j} \sum_{i=1}^{n_j} \b{\phi}(\b{x}_i^{(j)}).
\end{align}

According to the representation theory \cite{alperin1993local}, any solution (direction) $\b{\phi}(\b{u}) \in \mathcal{H}$ must lie in the span of ``all'' the training vectors mapped to $\mathcal{H}$, i.e., $\b{\Phi}(\b{X}) = [\b{\phi}(\b{x}_1), \dots, \b{\phi}(\b{x}_n)] \in \mathbb{R}^{t\times n}$ (usually $t \gg d$). Note that $\mathcal{H}$ denotes the Hilbert space (feature space). Therefore, we can state that:
\begin{align}\label{equation_u_kernel_FDA}
\mathbb{R}^{t} \ni \b{\phi}(\b{u}) = \sum_{i=1}^n \theta_i\, \b{\phi}(\b{x}_i) = \b{\Phi}(\b{X})\, \b{\theta},
\end{align}
where $\mathbb{R}^n \ni \b{\theta} := [\theta_1, \dots, \theta_n]^\top$ is the unknown vector of coefficients, and $\b{\phi}(\b{u}) \in \mathbb{R}^t$ is the pulled Fisher direction to the feature space.
The pulled directions can be put together in $\mathbb{R}^{t \times p} \ni \b{\Phi}(\b{U}) := [\b{\phi}(\b{u}_1), \dots, \b{\phi}(\b{u}_p)]$:
\begin{align}\label{equation_U_kernel_FDA}
\mathbb{R}^{t \times p} \ni \b{\Phi}(\b{U}) = \b{\Phi}(\b{X})\, \b{\Theta},
\end{align}
where $\b{\Theta} := [\b{\theta}_1, \dots, \b{\theta}_p] \in \mathbb{R}^{n \times p}$.

The $d_B$ in the feature space is:
\begin{align}
\mathbb{R} \ni d_B &:= \b{\phi}(\b{u})^\top \b{\Phi}(\b{S}_B)\, \b{\phi}(\b{u}) \\
&\overset{(a)}{=} \b{\theta}^\top \b{\Phi}(\b{X})^\top \big(\b{\phi}(\b{\mu}_1) - \b{\phi}(\b{\mu}_2)\big) \nonumber \\
&~~~~~~~~~~~~~~~~~~~ \big(\b{\phi}(\b{\mu}_1) - \b{\phi}(\b{\mu}_2)\big)^\top \b{\Phi}(\b{X})\, \b{\theta}, \label{equation_kernel_Fisher_twoClass_d_B_1}
\end{align}
where $(a)$ is because of Eqs. (\ref{equation_kernel_Fisher_twoClass_between_scatter}). and (\ref{equation_u_kernel_FDA}).

For the $j$-th class (here $j \in \{1,2\}$), we have:
\begin{align}
&\b{\theta}^\top \b{\Phi}(\b{X})^\top \b{\phi}(\b{\mu}_j) \overset{(\ref{equation_u_kernel_FDA})}{=} \sum_{i=1}^n \theta_i\, \b{\phi}(\b{x}_i)^\top \b{\phi}(\b{\mu}_j) \nonumber \\
&\overset{(\ref{equation_mean_of_class_inFeatureSpace})}{=} \frac{1}{n_j} \sum_{i=1}^n \sum_{k=1}^{n_j} \theta_i\, \b{\phi}(\b{x}_i)^\top \b{\phi}(\b{x}_k^{(j)}) \nonumber \\
&\overset{(\ref{equation_kernel_scalar})}{=} \frac{1}{n_j} \sum_{i=1}^n \sum_{k=1}^{n_j} \theta_i\, k(\b{x}_i, \b{x}_k^{(j)}) = \b{\theta}^\top \b{m}_j, \label{equation_kernel_Fisher_twoClass_theta_m}
\end{align}
where $\b{m}_j \in \mathbb{R}^n$ whose $i$-th entry is:
\begin{align}
\b{m}_j(i) := \frac{1}{n_j} \sum_{k=1}^{n_j} k(\b{x}_i, \b{x}_k^{(j)}).
\end{align}
Hence, Eq. (\ref{equation_kernel_Fisher_twoClass_d_B_1}) becomes:
\begin{align}
d_B \overset{(\ref{equation_kernel_Fisher_twoClass_theta_m})}{=} \b{\theta}^\top (\b{m}_1 - \b{m}_2) (\b{m}_1 - \b{m}_2)^\top \b{\theta} = \b{\theta}^\top \b{M} \b{\theta},
\end{align}
where:
\begin{align}\label{equation_kernel_Fisher_twoClass_M}
\mathbb{R}^{n \times n} \ni \b{M} := (\b{m}_1 - \b{m}_2) (\b{m}_1 - \b{m}_2)^\top,
\end{align}
is the \textit{between-scatter} in kernel FDA.
Hence, the Eq. (\ref{equation_kernel_Fisher_twoClass_d_B_1}) becomes:
\begin{align}\label{equation_kernel_Fisher_twoClass_d_B_2_oneDimensional}
d_B = \b{\phi}(\b{u})^\top \b{\Phi}(\b{S}_B)\, \b{\phi}(\b{u}) = \b{\theta}^\top \b{M} \b{\theta}.
\end{align}

The Eq. (\ref{equation_within_scatter_multipleClasses}) in the feature space is:
\begin{align}\label{equation_kernel_Fisher_twoClass_within_scatter}
&\mathbb{R}^{t \times t} \ni \b{\Phi}(\b{S}_W) := \nonumber \\
&~~~~~~~~~~~ \sum_{j=1}^c \sum_{i=1}^{n_j} \big(\b{\phi}(\b{x}_i^{(j)}) - \b{\phi}(\b{\mu}_j)\big) \big(\b{\phi}(\b{x}_i^{(j)}) - \b{\phi}(\b{\mu}_j)\big)^\top.
\end{align}

The $d_W$ in the feature space is:
\begin{align*}
&\mathbb{R} \ni d_W := \b{\phi}(\b{u})^\top \b{\Phi}(\b{S}_W)\, \b{\phi}(\b{u}) \\
&\overset{(a)}{=} \Big( \sum_{\ell=1}^n \theta_\ell\, \b{\phi}(\b{x}_\ell)^\top \Big) \Big( \sum_{j=1}^c \sum_{i=1}^{n_j} \big(\b{\phi}(\b{x}_i^{(j)}) - \b{\phi}(\b{\mu}_j)\big) \nonumber \\
&~~~~~~~~~~~~~~~~~~~ \big(\b{\phi}(\b{x}_i^{(j)}) - \b{\phi}(\b{\mu}_j)\big)^\top \Big) \Big( \sum_{k=1}^n \theta_k\, \b{\phi}(\b{x}_k) \Big) \\
&= \sum_{j=1}^c \sum_{\ell=1}^n \sum_{i=1}^{n_j} \sum_{k=1}^n \Big( \theta_\ell\, \b{\phi}(\b{x}_\ell)^\top \big(\b{\phi}(\b{x}_i^{(j)}) - \b{\phi}(\b{\mu}_j)\big) \\
&~~~~~~~~~~~~~~~~~~~ \big(\b{\phi}(\b{x}_i^{(j)}) - \b{\phi}(\b{\mu}_j)\big)^\top \theta_k\, \b{\phi}(\b{x}_k) \Big) 
\end{align*}

\begin{align*}
&\overset{(\ref{equation_mean_of_class_inFeatureSpace})}{=} \sum_{j=1}^c \sum_{\ell=1}^n \sum_{i=1}^{n_j} \sum_{k=1}^n \\
&~~~~~~~~~~~~~~~~~~~ \Big( \theta_\ell\, \b{\phi}(\b{x}_\ell)^\top \big(\b{\phi}(\b{x}_i^{(j)}) - \frac{1}{n_j} \sum_{e=1}^{n_j} \b{\phi}(\b{x}_e^{(j)})\big) \\
&~~~~~~~~~~~~~~~~~~~ \big(\b{\phi}(\b{x}_i^{(j)}) - \frac{1}{n_j} \sum_{z=1}^{n_j} \b{\phi}(\b{x}_z^{(j)})\big)^\top \theta_k\, \b{\phi}(\b{x}_k) \Big) 
\end{align*}

\begin{align*}
&\overset{(\ref{equation_kernel_scalar})}{=} \sum_{j=1}^c \sum_{\ell=1}^n \sum_{i=1}^{n_j} \sum_{k=1}^n \\
&~~~~~~~~~~~~~~~~~~~ \Big( \theta_\ell\, k(\b{x}_\ell, \b{x}_i^{(j)}) - \frac{1}{n_j} \sum_{e=1}^{n_j} \theta_\ell\, k(\b{x}_\ell, \b{x}_e^{(j)}) \Big) \\
&~~~~~~~~~~~~~~~~~~~ \Big( \theta_k\, k(\b{x}_i^{(j)}, \b{x}_k) - \frac{1}{n_j} \sum_{z=1}^{n_j} \theta_k\, k(\b{x}_z^{(j)}, \b{x}_k) \Big) 
\end{align*}

\begin{align*}
&\overset{(b)}{=} \sum_{j=1}^c \sum_{\ell=1}^n \sum_{i=1}^{n_j} \sum_{k=1}^n \\
&~~~~~~~~~~~~~~~~~~~ \Big( \theta_\ell\, k(\b{x}_\ell, \b{x}_i^{(j)}) - \frac{1}{n_j} \sum_{e=1}^{n_j} \theta_\ell\, k(\b{x}_\ell, \b{x}_e^{(j)}) \Big) \\
&~~~~~~~~~~~~~~~~~~~ \Big( \theta_k\, k(\b{x}_k, \b{x}_i^{(j)}) - \frac{1}{n_j} \sum_{z=1}^{n_j} \theta_k\, k(\b{x}_k, \b{x}_z^{(j)}) \Big) 
\end{align*}

\begin{align*}
&= \sum_{j=1}^c \sum_{\ell=1}^n \sum_{i=1}^{n_j} \sum_{k=1}^n \\
&\Big( \theta_\ell\, \theta_k\, k(\b{x}_\ell, \b{x}_i^{(j)})\, k(\b{x}_k, \b{x}_i^{(j)}) \\
&- \frac{2\,\theta_\ell\, \theta_k}{n_j} \sum_{z=1}^{n_j} k(\b{x}_\ell, \b{x}_i^{(j)})\, k(\b{x}_k, \b{x}_z^{(j)}) \\
& + \frac{\theta_\ell\, \theta_k}{n_j^2} \sum_{e=1}^{n_j} \sum_{z=1}^{n_j} k(\b{x}_\ell, \b{x}_e^{(j)})\, k(\b{x}_k, \b{x}_z^{(j)}) \Big)
\end{align*}

\begin{align*}
&= \sum_{j=1}^c \sum_{\ell=1}^n \sum_{i=1}^{n_j} \sum_{k=1}^n \\
&\Big( \theta_\ell\, \theta_k\, k(\b{x}_\ell, \b{x}_i^{(j)})\, k(\b{x}_k, \b{x}_i^{(j)}) \\
&- \frac{\theta_\ell\, \theta_k}{n_j} \sum_{z=1}^{n_j} k(\b{x}_\ell, \b{x}_i^{(j)})\, k(\b{x}_k, \b{x}_z^{(j)}) \Big) 
\end{align*}

\begin{align*}
&= \sum_{j=1}^c \bigg( \sum_{\ell=1}^n \sum_{i=1}^{n_j} \sum_{k=1}^n \Big( \theta_\ell\, \theta_k\, k(\b{x}_\ell, \b{x}_i^{(j)})\, k(\b{x}_k, \b{x}_i^{(j)}) \Big) \\
&- \sum_{\ell=1}^n \sum_{i=1}^{n_j} \sum_{k=1}^n \Big( \frac{\theta_\ell\, \theta_k}{n_j} \sum_{z=1}^{n_j} k(\b{x}_\ell, \b{x}_i^{(j)})\, k(\b{x}_k, \b{x}_z^{(j)}) \Big) \bigg) 
\end{align*}

\begin{align*}
&\overset{(c)}{=} \sum_{j=1}^c \big( \b{\theta}^\top \b{K}_j \b{K}_j^\top \b{\theta} - \b{\theta}^\top \b{K}_j \frac{1}{n_j} \b{1}\b{1}^\top \b{K}_j^\top \b{\theta} \big) \\
&= \sum_{j=1}^c \b{\theta}^\top \b{K}_j \big( \b{I} - \frac{1}{n_j} \b{1}\b{1}^\top \big) \b{K}_j^\top \b{\theta} \\
&\overset{(d)}{=} \sum_{j=1}^c \b{\theta}^\top \b{K}_j \b{H}_j \b{K}_j^\top \b{\theta} = \b{\theta}^\top \Big( \sum_{j=1}^c \b{K}_j \b{H}_j \b{K}_j^\top \Big) \b{\theta},
\end{align*}
where $(a)$ is because of Eqs. (\ref{equation_kernel_Fisher_twoClass_within_scatter}) and (\ref{equation_u_kernel_FDA}), $(b)$ is because $k(\b{x}_1, \b{x}_2) = k(\b{x}_2, \b{x}_1) \in \mathbb{R}$, and $(c)$ is because $\b{K}_j \in \mathbb{R}^{n \times n_j}$ is the kernel matrix of the whole training data and the training data of the $j$-th class. The $(a,b)$-th element of $\b{K}_j$ is:
\begin{align}
\b{K}_j(a,b) := k(\b{x}_a, \b{x}_b^{(j)}).
\end{align}
The $(d)$ is because:
\begin{align}
\mathbb{R}^{n_j \times n_j} \ni \b{H}_j := \b{I} - \frac{1}{n_j} \b{1}\b{1}^\top,
\end{align}
is the \textit{centering matrix} (see Appendix A in \cite{ghojogh2019unsupervised}).

We define:
\begin{align}\label{equation_kernel_Fisher_N}
\mathbb{R}^{n \times n} \ni \b{N} := \sum_{j=1}^c \b{K}_j \b{H}_j \b{K}_j^\top,
\end{align}
as the \textit{within-scatter} in kernel FDA.
Hence, the $d_W$ becomes:
\begin{align}\label{equation_kernel_Fisher_twoClass_d_W_2}
d_W = \b{\phi}(\b{u})^\top \b{\Phi}(\b{S}_W)\, \b{\phi}(\b{u}) = \b{\theta}^\top \b{N} \b{\theta}.
\end{align}

The kernel Fisher criterion is:
\begin{align}\label{equation_kernel_Fisher_criterion}
f(\b{\theta}) := \frac{d_B(\b{\theta})}{d_W(\b{\theta})} = \frac{\b{\phi}(\b{u})^\top \b{\Phi}(\b{S}_B)\, \b{\phi}(\b{u})}{\b{\phi}(\b{u})^\top \b{\Phi}(\b{S}_W)\, \b{\phi}(\b{u})} = \frac{\b{\theta}^\top \b{M} \b{\theta}}{\b{\theta}^\top \b{N} \b{\theta}},
\end{align}
where the $\b{\theta} \in \mathbb{R}^n$ is the \textit{kernel Fisher direction}. 

Similar to the solution of Eq. (\ref{equation_Fisher_optimization_oneDirection_1}), the solution to maximization of Eq. (\ref{equation_kernel_Fisher_criterion}) is:
\begin{align}
\b{M}\b{\theta} = \lambda\, \b{N} \b{\theta},
\end{align}
which is a generalized eigenvalue problem $(\b{M}, \b{N})$ according to \cite{ghojogh2019eigenvalue}. The $\b{\theta}$ is the eigenvector with the largest eigenvalue (because the optimization is maximization) and the $\lambda$ is the corresponding eigenvalue.
The $\b{\theta}$ is the \textit{kernel Fisher direction} or \textit{kernel Fisher axis}.

Again, one possible solution to the generalized eigenvalue problem $(\b{M}, \b{N})$ is \cite{ghojogh2019eigenvalue}:
\begin{align}
\b{\theta} = \textbf{eig}(\b{N}^{-1} \b{M}), \label{equation_kernel_Fisher_solution_oneDirection}
\end{align}
or \cite{ghojogh2019eigenvalue}:
\begin{align}
\b{\theta} = \textbf{eig}((\b{N} + \varepsilon \b{I})^{-1} \b{M}),
\end{align}
where $\textbf{eig}(.)$ denotes the eigenvector of the matrix with the largest eigenvalue.

The projection and reconstruction of the training data point $\b{x}_i$ and the out-of-sample data point $\b{x}_t$ are:
\begin{align}
&\mathbb{R} \ni \phi(\widetilde{\b{x}}_i) = \b{\phi}(\b{u})^\top \b{\phi}(\b{x}_i) \overset{(\ref{equation_u_kernel_FDA})}{=} \b{\theta}^\top \b{\Phi}(\b{X})^\top \b{\phi}(\b{x}_i) \nonumber \\
&~~~~~~~~~ = \b{\theta}^\top \b{k}(\b{X}, \b{x}_i), \\
&\mathbb{R}^t \ni \b{\phi}(\widehat{\b{x}}_i) = \b{\phi}(\b{u}) \b{\phi}(\b{u})^\top \b{\phi}(\b{x}_i) \nonumber \\
&~~~~~~~~~ \overset{(\ref{equation_u_kernel_FDA})}{=} \b{\Phi}(\b{X})\, \b{\theta} \b{\theta}^\top \b{k}(\b{X}, \b{x}_i), \\
&\mathbb{R} \ni \phi(\widetilde{\b{x}}_t) = \b{\theta}^\top \b{k}(\b{X}, \b{x}_t), \\
&\mathbb{R}^t \ni \b{\phi}(\widehat{\b{x}}_t) = \b{\Phi}(\b{X})\, \b{\theta} \b{\theta}^\top \b{k}(\b{X}, \b{x}_t).
\end{align}
However, in reconstruction expressions, the $\b{\Phi}(\b{X})$ is not necessarily available; therefore, in kernel FDA, similar to kernel PCA \cite{ghojogh2019unsupervised}, \textit{reconstruction cannot be done.}
For the whole training and out-of-sample data, the projections are:
\begin{align}
&\mathbb{R}^{1 \times n} \ni \b{\Phi}(\widetilde{\b{X}}) = \b{\theta}^\top \b{K}(\b{X}, \b{X}), \\
&\mathbb{R}^{1 \times n_t} \ni \b{\Phi}(\widetilde{\b{X}}_t) = \b{\theta}^\top \b{K}(\b{X}, \b{X}_t).
\end{align}

\subsubsection{Scatters in Multi-Class Case: Variant 1}

In multi-class case for kernel FDA, the \textit{within-scatter} is the same as in the two-class case, which is Eq. (\ref{equation_kernel_Fisher_N}) and $d_W$ is also Eq. (\ref{equation_kernel_Fisher_twoClass_d_W_2}). However, the between-scatter is different.
The between-scatter, Eq. (\ref{equation_between_scatter_multipleClasses}), in the feature space is:
\begin{align}\label{equation_kernel_Fisher_MultiClass_between_scatter}
&\mathbb{R}^{t \times t} \ni \b{\Phi}(\b{S}_B) := \nonumber \\
&~~~~~~~~~~~ \sum_{j=1}^c \big(\b{\phi}(\b{\mu}_j) - \b{\phi}(\b{\mu})\big) \big(\b{\phi}(\b{\mu}_j) - \b{\phi}(\b{\mu})\big)^\top,
\end{align}
where the total mean in the feature space is:
\begin{align}\label{equation_total_mean_inFeatureSpace}
\mathbb{R}^{t} \ni \b{\phi}(\b{\mu}) := \frac{1}{\sum_{k=1}^c n_k} \sum_{j=1}^c n_j\, \b{\phi}(\b{\mu}_j) = \frac{1}{n} \sum_{i=1}^n \b{\phi}(\b{x}_i),
\end{align}

The $d_B$ in the feature space is:
\begin{align}
\mathbb{R} \ni d_B &:= \b{\phi}(\b{u})^\top \b{\Phi}(\b{S}_B)\, \b{\phi}(\b{u}) \nonumber \\
&\overset{(a)}{=} \sum_{j=1}^c \b{\theta}^\top \b{\Phi}(\b{X})^\top \big(\b{\phi}(\b{\mu}_j) - \b{\phi}(\b{\mu})\big) \nonumber \\
&~~~~~~~~~~~~~~~~~~~ \big(\b{\phi}(\b{\mu}_j) - \b{\phi}(\b{\mu})\big)^\top \b{\Phi}(\b{X})\, \b{\theta}, \label{equation_kernel_Fisher_MultiClass_d_B_1} 
\end{align}
where $(a)$ is because of Eqs. (\ref{equation_kernel_Fisher_MultiClass_between_scatter}) and (\ref{equation_u_kernel_FDA}).
We have:
\begin{align}
&\b{\theta}^\top \b{\Phi}(\b{X})^\top \b{\phi}(\b{\mu}) \overset{(\ref{equation_u_kernel_FDA})}{=} \sum_{i=1}^n \theta_i\, \b{\phi}(\b{x}_i)^\top \b{\phi}(\b{\mu}) \nonumber \\
&\overset{(\ref{equation_total_mean_inFeatureSpace})}{=} \frac{1}{n} \sum_{i=1}^n \sum_{k=1}^{n} \theta_i\, \b{\phi}(\b{x}_i)^\top \b{\phi}(\b{x}_k) \nonumber \\
&\overset{(\ref{equation_kernel_scalar})}{=} \frac{1}{n} \sum_{i=1}^n \sum_{k=1}^{n} \theta_i\, k(\b{x}_i, \b{x}_k) = \b{\theta}^\top \b{m}_*, \label{equation_kernel_Fisher_MultiClass_theta_m}
\end{align}
where $\b{m}_* \in \mathbb{R}^n$ whose $i$-th entry is:
\begin{align}\label{equation_kernel_Fisher_MultiClass_m_star}
\b{m}_*(i) := \frac{1}{n} \sum_{k=1}^{n} k(\b{x}_i, \b{x}_k).
\end{align}

According to Eqs. (\ref{equation_kernel_Fisher_twoClass_theta_m}) and (\ref{equation_kernel_Fisher_MultiClass_theta_m}), the  Eq. (\ref{equation_kernel_Fisher_MultiClass_d_B_1}) becomes:
\begin{align}
d_B = \b{\theta}^\top \sum_{j=1}^c (\b{m}_j - \b{m}_*) (\b{m}_j - \b{m}_*)^\top \b{\theta} = \b{\theta}^\top \b{M} \b{\theta},
\end{align}
where:
\begin{align}\label{equation_kernel_Fisher_MultiClass_M}
\mathbb{R}^{n \times n} \ni \b{M} := \sum_{j=1}^c (\b{m}_j - \b{m}_*) (\b{m}_j - \b{m}_*)^\top,
\end{align}
is the \textit{between-scatter} in kernel FDA.
Similar to Eq. (\ref{equation_between_scatter_multipleClasses_2}), some researches consider the following instead:
\begin{align}\label{equation_kernel_Fisher_MultiClass_M_2}
\mathbb{R}^{n \times n} \ni \b{M} := \sum_{j=1}^c n_j\, (\b{m}_j - \b{m}_*) (\b{m}_j - \b{m}_*)^\top.
\end{align}

Hence, the Eq. (\ref{equation_kernel_Fisher_MultiClass_d_B_1}) becomes:
\begin{align}\label{equation_kernel_Fisher_MultiClass_d_B_2_multiDimensional}
d_B = \b{\phi}(\b{u})^\top \b{\Phi}(\b{S}_B)\, \b{\phi}(\b{u}) = \b{\theta}^\top \b{M} \b{\theta},
\end{align}
where $\b{M}$ here is Eq. (\ref{equation_kernel_Fisher_MultiClass_M}) or (\ref{equation_kernel_Fisher_MultiClass_M_2}).

The Fisher direction is again Eq. (\ref{equation_kernel_Fisher_criterion}) and the solution is again the generalized eigenvalue problem $(\b{M}, \b{N})$ according to \cite{ghojogh2019eigenvalue}. 

\subsubsection{Scatters in Multi-Class Case: Variant 2}

Again, in the second version of multi-class case for kernel FDA, the \textit{within-scatter} is the same as in the two-class case, which is Eq. (\ref{equation_kernel_Fisher_N}) and $d_W$ is also Eq. (\ref{equation_kernel_Fisher_twoClass_d_W_2}).

For the between scatter in in the second version, we start with the Eqs. (\ref{equation_Fisher_criterion_variant2_oneDimensional}) and (\ref{equation_optimization_FDA_with_S_T_oneDirection}). We kernelize the objective function of the Eq. (\ref{equation_optimization_FDA_with_S_T_oneDirection}):
\begin{align}\label{equation_kernel_fisher_d_T}
d_T := \b{\phi}(\b{u})^\top \b{\Phi}(\b{S}_T)\, \b{\phi}(\b{u}),
\end{align}
where total-scatter, Eq. (\ref{equation_fisher_total_scatter}), is pulled as:
\begin{align}\label{equation_kernel_Fisher_multiClass_total_scatter}
&\mathbb{R}^{t \times t} \ni \b{\Phi}(\b{S}_T) := \nonumber \\
&~~~~~~~~~~~~~~~ \sum_{k=1}^n \big(\b{\phi}(\b{x}_k) - \b{\phi}(\b{\mu})\big) \big(\b{\phi}(\b{x}_k) - \b{\phi}(\b{\mu})\big)^\top.
\end{align}
According to Eqs. (\ref{equation_u_kernel_FDA}), (\ref{equation_kernel_fisher_d_T}), and (\ref{equation_kernel_Fisher_multiClass_total_scatter}), we have:
\begin{align*}
&d_T = \sum_{k=1}^n \b{\theta}^\top \b{\Phi}(\b{X})^\top \big(\b{\phi}(\b{x}_k) - \b{\phi}(\b{\mu})\big)\\
&~~~~~~~~~~~~~~~~~~~~~~~~ \big(\b{\phi}(\b{x}_k) - \b{\phi}(\b{\mu})\big)^\top \b{\Phi}(\b{X})\, \b{\theta}.
\end{align*}
According to Eq. (\ref{equation_kernel_Fisher_MultiClass_theta_m}), we have:
\begin{align}
\b{\theta}^\top \b{\Phi}(\b{X})^\top \b{\phi}(\b{\mu}) = \b{\theta}^\top \b{m}_*,
\end{align}
where $\b{m}_*$ is Eq. (\ref{equation_kernel_Fisher_MultiClass_m_star}).
On the other hand, we have:
\begin{align}
\b{\theta}^\top \b{\Phi}(\b{X})^\top \b{\phi}(\b{x}_k) &\overset{(\ref{equation_u_kernel_FDA})}{=} \sum_{i=1}^n \theta_i\, \b{\phi}(\b{x}_i)^\top \b{\phi}(\b{x}_k) \nonumber \\
&\overset{(\ref{equation_kernel_scalar})}{=} \sum_{i=1}^n \theta_i\, k(\b{x}_i, \b{x}_k) = \b{\theta}^\top \b{g}_k,
\end{align}
where $\b{g}_k \in \mathbb{R}^n$ whose $i$-th entry is:
\begin{align}
\b{g}_k(i) := k(\b{x}_i, \b{x}_k).
\end{align}
Hence:
\begin{align}
d_T = \sum_{k=1}^n \b{\theta}^\top (\b{g}_k - \b{m}_*) (\b{g}_k - \b{m}_*)^\top \b{\theta} = \b{\theta}^\top \b{G}\, \b{\theta},
\end{align}
where:
\begin{align}
\mathbb{R}^{n \times n} \ni \b{G} := \sum_{k=1}^n (\b{g}_k - \b{m}_*) (\b{g}_k - \b{m}_*)^\top.
\end{align}
The denominator of the Fisher criterion in the feature space is again the Eq. (\ref{equation_kernel_Fisher_twoClass_d_W_2}). 

The optimization will be similar to Eq. (\ref{equation_optimization_FDA_with_S_T_oneDirection}) but in the feature space:
\begin{equation}\label{equation_optimization_kernel_FDA_with_S_T_oneDirection}
\begin{aligned}
& \underset{\b{\theta}}{\text{maximize}}
& & \b{\theta}^\top \b{G}\, \b{\theta} \\
& \text{subject to}
& & \b{\theta}^\top \b{N}\, \b{\theta} = 1,
\end{aligned}
\end{equation}
whose solution is similarly obtained as:
\begin{align}
\b{G}\,\b{\theta} = \lambda\, \b{N}\, \b{\theta},
\end{align}
which is a generalized eigenvalue problem $(\b{G}, \b{N})$ according to \cite{ghojogh2019eigenvalue}.

\begin{figure*}[!t]
\centering
\includegraphics[width=6in]{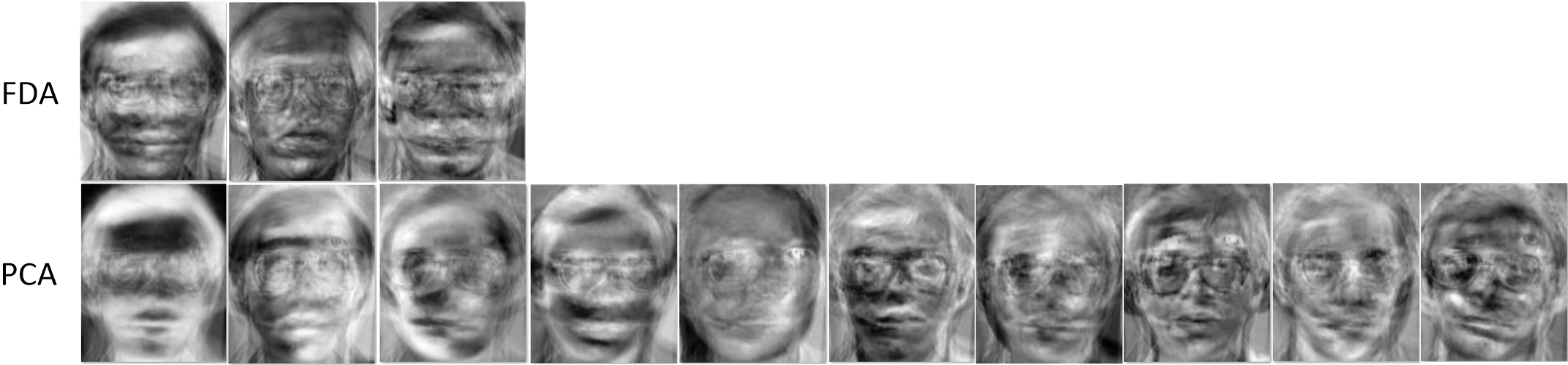}
\caption{The projection directions (ghost faces) of FDA and PCA for the first four classes of facial AT\&T dataset.}
\label{figure_projection_directions}
\end{figure*}

\subsection{Multi-dimensional Subspace}

In the previous section, the one-dimensional kernel Fisher subspace was discussed. In multi-dimensional kernel Fisher subspace, the within- and between-scatters are the same but the Fisher criterion is different.
According to Eq. (\ref{equation_U_kernel_FDA}), the $d_B$ and $d_W$ are:
\begin{align}
&d_B = \textbf{tr}\big(\b{\phi}(\b{U})^\top \b{\Phi}(\b{S}_B)\, \b{\phi}(\b{U})\big) = \textbf{tr}(\b{\Theta}^\top \b{M} \b{\Theta}), \\
&d_W = \textbf{tr}\big(\b{\phi}(\b{U})^\top \b{\Phi}(\b{S}_W)\, \b{\phi}(\b{U})\big) = \textbf{tr}(\b{\Theta}^\top \b{N} \b{\Theta}),
\end{align}
where $\mathbb{R}^{n \times p} \ni \b{\Theta} = [\b{\theta}_1, \dots, \b{\theta}_p]$ and $\b{M} \in \mathbb{R}^{n \times n}$ and $\b{N} \in \mathbb{R}^{n \times n}$ are the between- and within-scatters, respectively, determined for either two-class or multi-class case.

The Fisher criterion becomes:
\begin{align}\label{equation_kernel_Fisher_criterion_multiDimensional}
f(\b{\Theta}) &:= \frac{d_B(\b{\Theta})}{d_W(\b{\Theta})} = \frac{\textbf{tr}\big(\b{\phi}(\b{U})^\top \b{\Phi}(\b{S}_B)\, \b{\phi}(\b{U})\big)}{\textbf{tr}\big(\b{\phi}(\b{U})^\top \b{\Phi}(\b{S}_W)\, \b{\phi}(\b{U})\big)} \nonumber \\
&= \frac{\textbf{tr}(\b{\Theta}^\top \b{M} \b{\Theta})}{\textbf{tr}(\b{\Theta}^\top \b{N} \b{\Theta})},
\end{align}
where the columns of $\b{\Theta}$ are the \textit{kernel Fisher directions}.

Similar to Eq. (\ref{equation_Fisher_optimization_SeveralDirections_1}), the solution to maximization of this criterion is:
\begin{align}
\b{M}\b{\Theta} = \b{N} \b{\Theta} \b{\Lambda},
\end{align}
which is the generalized eigenvalue problem $(\b{M}, \b{N})$ according to \cite{ghojogh2019eigenvalue}. The columns of $\b{\Theta}$ are the eigenvectors sorted from the largest to smallest eigenvalues (because the optimization is maximization) and the diagonal entries of $\b{\Lambda}$ are the corresponding eigenvalues.

Again, we can have another variant of kernel FDA for the multi-dimensional sub-space where the optimization is (similar to Eq. (\ref{equation_optimization_kernel_FDA_with_S_T_oneDirection})):
\begin{equation}\label{equation_optimization_kernel_FDA_with_S_T_multiDirections}
\begin{aligned}
& \underset{\b{\Theta}}{\text{maximize}}
& & \textbf{tr} (\b{\Theta}^\top \b{G}\, \b{\Theta}) \\
& \text{subject to}
& & \b{\Theta}^\top \b{N}\, \b{\Theta} = \b{I},
\end{aligned}
\end{equation}
whose solution is similarly obtained as:
\begin{align}
\b{G}\b{\Theta} = \b{N} \b{\Theta} \b{\Lambda},
\end{align}
which is a generalized eigenvalue problem $(\b{G}, \b{N})$ according to \cite{ghojogh2019eigenvalue}.

As mentioned before, in kernel FDA, we do not have reconstruction. The projection of the training data point $\b{x}_i$ and the out-of-sample data point $\b{x}_t$ are:
\begin{align}
&\mathbb{R}^p \ni \b{\phi}(\widetilde{\b{x}}_i) = \b{\Phi}(\b{U})^\top \b{\phi}(\b{x}_i) \overset{(\ref{equation_U_kernel_FDA})}{=} \b{\Theta}^\top \b{\Phi}(\b{X})^\top \b{\phi}(\b{x}_i) \nonumber \\
&~~~~~~~~~ = \b{\Theta}^\top \b{k}(\b{X}, \b{x}_i), \\
&\mathbb{R}^p \ni \b{\phi}(\widetilde{\b{x}}_t) = \b{\Theta}^\top \b{k}(\b{X}, \b{x}_t).
\end{align}
For the whole training and out-of-sample data, the projections are:
\begin{align}
&\mathbb{R}^{p \times n} \ni \b{\Phi}(\widetilde{\b{X}}) = \b{\Theta}^\top \b{K}(\b{X}, \b{X}), \\
&\mathbb{R}^{p \times n_t} \ni \b{\Phi}(\widetilde{\b{X}}_t) = \b{\Theta}^\top \b{K}(\b{X}, \b{X}_t).
\end{align}

\begin{figure*}[!t]
\centering
\includegraphics[width=6.5in]{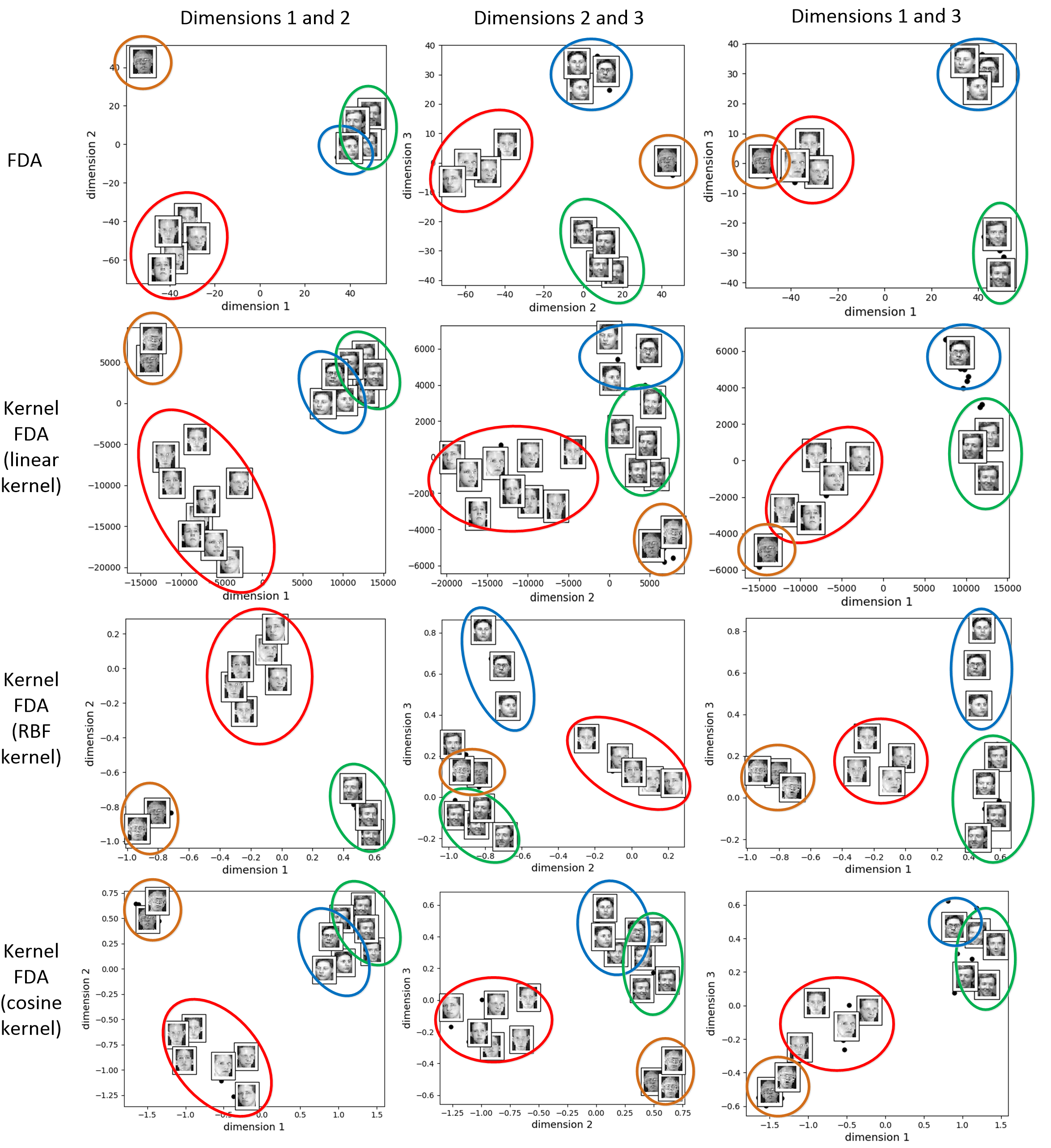}
\caption{The projection of the first four classes of AT\&T dataset onto FDA and kernel FDA subspaces where the used kernels were linear, RBF, and cosine kernels.}
\label{figure_projection_FDA}
\end{figure*}

\begin{figure*}[!t]
\centering
\includegraphics[width=6.5in]{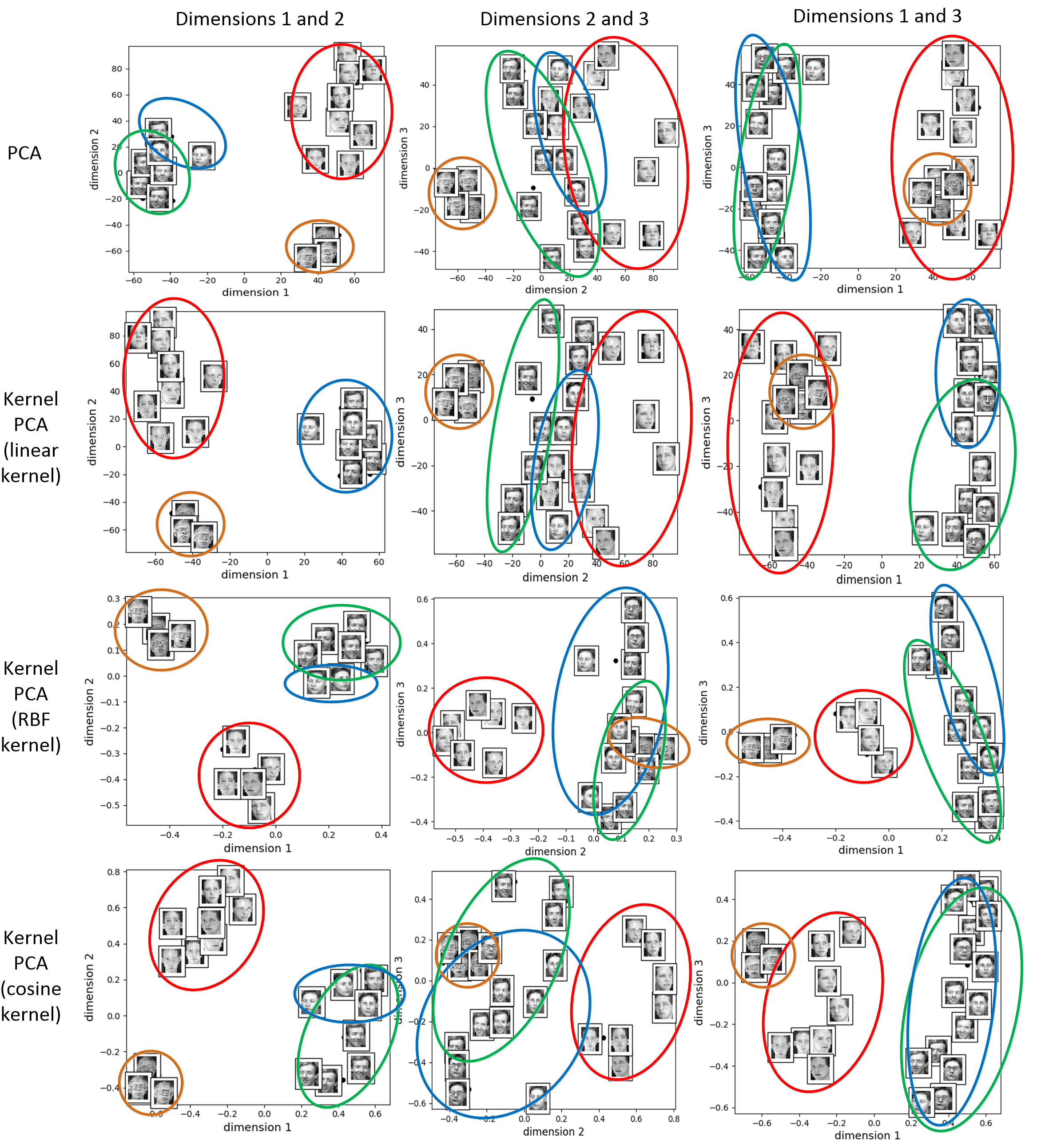}
\caption{The projection of the first four classes of AT\&T dataset onto PCA and kernel PCA subspaces where the used kernels were linear, RBF, and cosine kernels.}
\label{figure_projection_PCA}
\end{figure*}

\subsection{Discussion on Dimensionality of the Kernel Fisher Subspace}

According to Eq. (\ref{equation_kernel_Fisher_N}), the rank of the $\b{N}$ is at most $\min(n, c)$ because the matrix is $n \times n$ and its calculation includes $c$ iterations. Hence, the rank of $\b{N}^{-1}$ is also at most $\min(n, c)$.
According to Eq. (\ref{equation_kernel_Fisher_MultiClass_M_2}), the rank of the $\b{M}$ is at most $\min(n, c-1)$ because the matrix is $n \times n$, we have $c$ iterations in its calculation, and $-1$ is because of subtracting the mean (refer to the explanation in Section \ref{section_Fisher_dimensionality}). 

In Eq. (\ref{equation_kernel_Fisher_solution_oneDirection}), we have $\b{N}^{-1} \b{M}$ whose rank is:
\begin{align}
&\textbf{rank}(\b{N}^{-1} \b{M}) \leq \min\big(\textbf{rank}(\b{N}^{-1}), \textbf{rank}(\b{M})\big) \nonumber \\
&\leq \min\big(\min(n, c), \min(n, c-1)\big) \nonumber \\
&= \min(n, c, c-1) \overset{(a)}{=} c-1,
\end{align}
where $(a)$ is because we usually have $c < n$.
Therefore, the rank of $\b{N}^{-1} \b{M}$ is limited because of the rank of $\b{M}$ which is at most $c-1$. 

According to Eq. (\ref{equation_kernel_Fisher_solution_oneDirection}), the $c-1$ leading eigenvalues will be valid and the rest are zero or very small. Therefore, the $p$, which is the dimensionality of the kernel Fisher subspace, is at most $c-1$. The $c-1$ leading eigenvectors are considered as the kernel Fisher directions and the rest of eigenvectors are invalid and ignored.

\section{Fisherfaces}

This section introduces one of the most fundamental applications of PCA and its variants -- facial recognition.

\subsection{Projection Directions of Facial Images}

FDA and kernel FDA can be trained using images of diverse faces, to learn the most discriminative facial features, which separate the human subjects based on their facial pictures. Here, a facial dataset, i.e. the AT\&T (ORL) face dataset, is used to illustrate this concept.
Here, four different classes of the AT\&T (or ORL) facial dataset were used for training FDA, kernel FDA, PCA, and kernel PCA, where the used kernels were linear, Radial Basis Function (RBF), and cosine kernels. 
Since there are four classes, the number of FDA directions is three (because $c-1=3$).
The three FDA directions and the top ten PCA directions for the used dataset are shown in Fig. \ref{figure_projection_directions}. 
As can be seen, the projection directions of a facial dataset are some facial features which are like ghost faces. That is why the facial projection directions are also referred to as \textit{ghost faces}, while the ghost faces in FDA and PCA are also referred to as \textit{Fisherfaces} \cite{belhumeur1997eigenfaces,etemad1997discriminant,zhao1999subspace} and \textit{eigenfaces} \cite{turk1991eigenfaces,turk1991face}, respectively.
In Fig. \ref{figure_projection_directions}, the projection directions have captured different facial features, such as eyes, nose, cheeks, chin, lips, hair, and glasses, which discriminate the data with respect to the maximum variance in PCA, and maximum class separation and minimum within-class scatter in FDA. 
The extracted features by PCA and FDA are different because PCA extracts features for maximum variance between faces while FDA finds features which are different among the classes for their separation.
Figure \ref{figure_projection_directions} does not include projection directions for kernel FDA and kernel PCA because in kernel FDA, the projection directions are $n$-dimensional and not $d$-dimensional, and in kernel PCA, the projection directions are not available \cite{ghojogh2019unsupervised}.
Note that facial recognition using kernel FDA and kernel PCA are referred to as \textit{kernel Fisherfaces} \cite{yang2002kernel,liu2004improving} and \textit{kernel eigenfaces} \cite{yang2000face}, respectively.

\subsection{Projection of Facial Images}

The projection of the images onto FDA and kernel FDA subspaces are demonstrated in Fig. \ref{figure_projection_FDA}. The projection of the images using PCA and kernel PCA are also depicted in Fig. \ref{figure_projection_PCA}.
As can be seen, the FDA and kernel FDA subspaces have better separated the classes compared to the PCA and kernel PCA subspaces. This is because the FDA and kernel FDA make use of class labels in order to separate the classes in the subspace, while the PCA and kernel PCA only capture the variance (spread) of data regardless of class labels. 

\subsection{Reconstruction of Facial Images}

\begin{figure}[!t]
\centering
\includegraphics[width=3in]{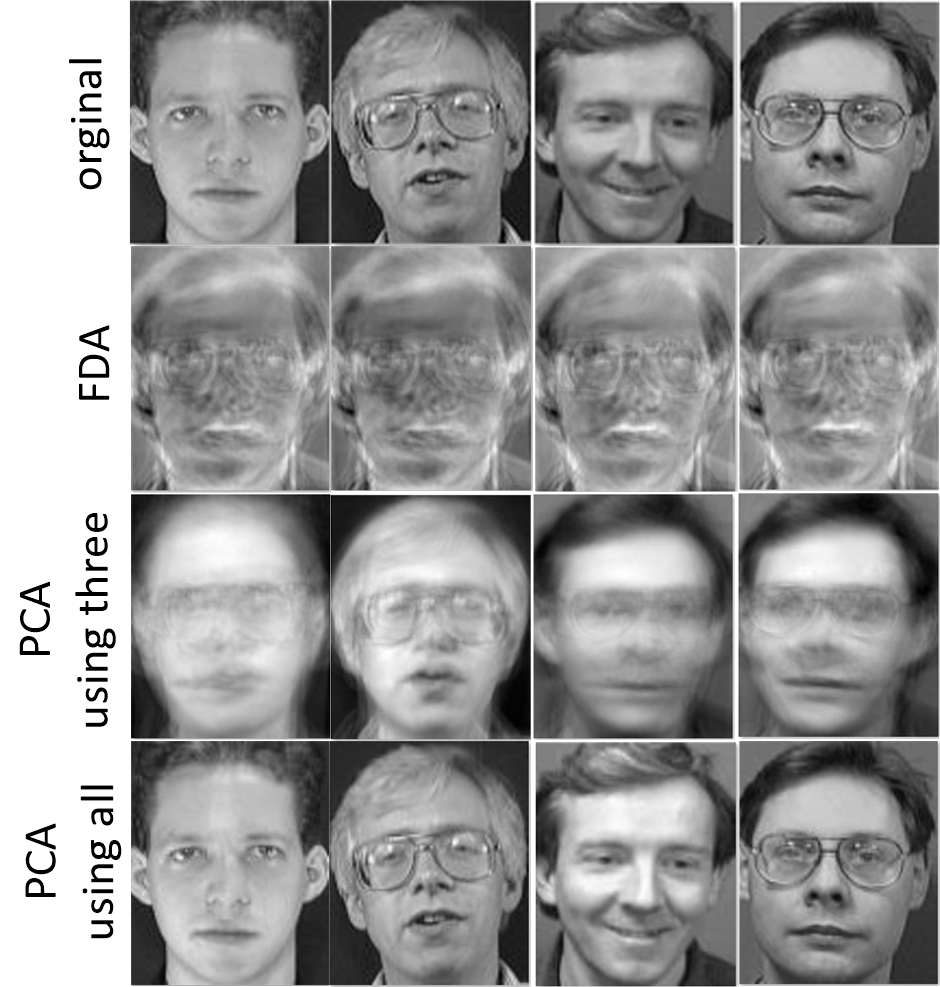}
\caption{The reconstruction of four sample faces of AT\&T datasets in FDA and PCA.}
\label{figure_reconstruction}
\end{figure}

Figure \ref{figure_reconstruction} illustrates the reconstruction of some training images. For reconstruction in this figure, FDA has used three projection directions (because $c-1=3$), PCA once has used the top three PCA directions and has also used the whole $d$ PCA directions. 
This figure demonstrates that the PCA reconstruction outperforms that of the FDA reconstruction. This makes sense because PCA is a linear method for reconstruction which has the least squared error \cite{ghojogh2019unsupervised}. However, the primary responsibility of FDA is not reconstruction, but separation of the classes. Thus, the FDA directions try to separate the classes as much as possible and do not \textit{necessarily} care for a good reconstruction. According to Fig. \ref{figure_PCA_FDA_comparison} , even in some datasets, the FDA direction may be orthogonal to the PCA optimal direction for reconstruction. 
It is noteworthy that reconstruction cannot be done in kernel FDA. However it can be done in FDA for the out-of-sample data (for the sake of brevity, a simulation is not provided).

\subsection{Out-of-sample Projection of Facial Images}

\begin{figure*}[!t]
\centering
\includegraphics[width=6.5in]{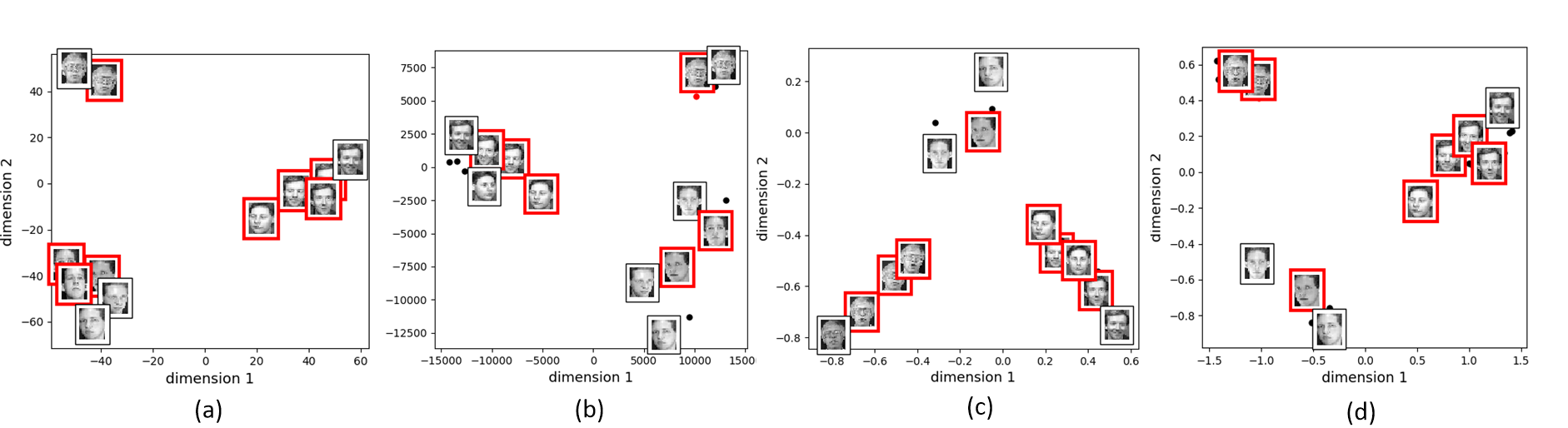}
\caption{The first two dimensions of the projection of both training and out-of-sample instances in the first four classes of AT\&T dataset onto subspaces of (a) FDA, (b) kernel FDA using linear kernel, (c) kernel FDA using RBF kernel, and (d) kernel FDA using cosine kernel.}
\label{figure_outOfSample}
\end{figure*}

The first six images of each of the first four subjects in the AT\&T dataset were taken as training images, while the rest of the images were used as test (out-of-sample) images. 
The projection of the training and the out-of-sample images onto FDA and kernel FDA (using linear, RBF, and cosine kernels) are shown in Fig. \ref{figure_outOfSample}.
This figure demonstrates that the projection of the out-of-sample images has been properly carried out in FDA and kernel FDA.
Therefore, FDA and kernel FDA can generalize well to the out-of-sample data that are not introduced to the model during training.

\section{Conclusion}

This paper was a tutorial paper introducing FDA and kernel FDA in detail. Various concepts about FDA, such as rank of scatters, dimensionality of the subspace, an example for interpretation, robust FDA, equivalency to LDA, and Fisher forest were explained and discussed. Both cases of two- and multi-classes were covered for FDA and kernel FDA. Finally, some simulations were performed to validate the theory in practice and compare to the unsupervised PCA method.

\section*{Acknowledgment}
The authors hugely thank Prof. Ali Ghodsi (see his great online courses \cite{web_data_visualization,web_classification}), Prof. Mu Zhu, Prof. Hoda Mohammadzade, and other professors whose courses have partly covered the materials mentioned in this tutorial paper.

\appendix

\section{Metric Learning}\label{section_appendix_metric_learning}

The general form of metric \cite{peltonen2004improved} is usually defined as a form similar to Mahalanobis distance \cite{mclachlan1999mahalanobis,de2000mahalanobis}. The metric is:
\begin{equation}
||\b{x}_i - \b{x}_j||_{\b{A}} := (\b{x}_i - \b{x}_j)^\top \b{A}\, (\b{x}_i - \b{x}_j),
\end{equation}
where:
\begin{align}\label{equation_appendix_matrix_A}
\b{A} = \b{U}\b{U}^\top \succeq 0,
\end{align}
to have a valid distance metric. 
Most of the metric learning algorithms \cite{kulis2013metric} are optimization problems where $\b{A}$ is unknown to make data points in same class (similar pairs) closer to each other, and points in different classes far apart from each other. We have:
\begin{align}
||\b{x}_i - \b{x}_j||_{\b{A}} &\overset{(\ref{equation_appendix_matrix_A})}{=} (\b{x}_i - \b{x}_j)^\top \b{U}\b{U}^\top\, (\b{x}_i - \b{x}_j) \label{equation_appendix_metrixLearning_projection1} \\
&= (\b{U}^\top \b{x}_i - \b{U}^\top \b{x}_j)^\top (\b{U}^\top \b{x}_i - \b{U}^\top \b{x}_j), \label{equation_appendix_metrixLearning_projection2}
\end{align}
so this metric is equivalent to projection of data with projection matrix $\b{U}$ and then using Euclidean distance in the embedded space \cite{peltonen2004improved}. Therefore, Metric learning can be considered as a feature extraction \cite{ghojogh2019feature} and manifold learning method \cite{alipanahi2008distance,globerson2006metric}.

\section{Rayleigh-Ritz Quotient}\label{section_appendix_rayleigh_ritz_quotient}

The \textit{Rayleigh-Ritz quotient} or \textit{Rayleigh quotient} is defined as \cite{parlett1998symmetric,croot2005rayleigh}:
\begin{align}\label{equation_rayleigh_ritz_quotient}
\mathbb{R} \ni R(\b{A}, \b{x}) := \frac{\b{x}^\top \b{A}\, \b{x}}{\b{x}^\top \b{x}},
\end{align}
where $\b{A}$ is a symmetric matrix and $\b{x}$ is a non-zero vector:
\begin{align}
\b{A} = \b{A}^\top, ~~ \b{x} \neq \b{0}.
\end{align}
One of the properties of the Rayleigh-Ritz quotient is:
\begin{align}\label{equation_rayleigh_ritz_quotient_scaling}
R(\b{A}, c\b{x}) = R(\b{A}, \b{x}),
\end{align}
where $c$ is a scalar. The proof is that:
\begin{align*}
R(\b{A}, c\b{x}) &= \frac{(c\b{x})^\top \b{A}\, c\b{x}}{(c\b{x})^\top c\b{x}} \overset{(a)}{=} \frac{c\b{x}^\top \b{A}\, c\b{x}}{c\b{x}^\top c\b{x}} \\
&\overset{(b)}{=} \frac{c^2}{c^2} \times \frac{\b{x}^\top \b{A}\, \b{x}}{\b{x}^\top \b{x}} \overset{(\ref{equation_rayleigh_ritz_quotient})}{=} R(\b{A}, \b{x}),
\end{align*}
where $(a)$ and $(b)$ are because $c$ is a scalar.

Because of the Eq. (\ref{equation_rayleigh_ritz_quotient_scaling}), the optimization of the Rayleigh-Ritz quotient has an equivalent \cite{croot2005rayleigh}:
\begin{align}
&\underset{\b{x}}{\text{minimize/maximize}} ~~ R(\b{A}, \b{x}) \overset{(a)}{\equiv} \nonumber \\
&\begin{aligned}
& \underset{\b{x}}{\text{minimize/maximize}}
& & R(\b{A}, \b{x}) \\
& \text{subject to}
& & ||\b{x}||_2 = 1,
\end{aligned} \overset{(b)}{\equiv} \nonumber \\
&\begin{aligned}\label{equation_rayleigh_ritz_quotient_optimization}
& \underset{\b{x}}{\text{minimize/maximize}}
& & \b{x}^\top \b{A}\, \b{x} \\
& \text{subject to}
& & ||\b{x}||_2 = 1,
\end{aligned}
\end{align}
where $(a)$ is because if we define $\b{y} := (1/||\b{x}||_2)\, \b{x}$, the Rayleigh-Ritz quotient is:
\begin{align}
R(\b{A}, \b{y}) = \frac{\b{y}^\top \b{A}\, \b{y}}{\b{y}^\top \b{y}} = \frac{1/||\b{x}||_2^2}{1/||\b{x}||_2^2} \times \frac{\b{x}^\top \b{A}\, \b{x}}{\b{x}^\top \b{x}} = R(\b{A}, \b{x}),
\end{align}
and:
\begin{align}
||\b{y}||_2^2 = \frac{1}{||\b{x}||_2^2} \times ||\b{x}||_2^2 = 1 \implies ||\b{y}||_2 = 1.
\end{align}
Thus, we have $R(\b{A}, \b{y})$ subject to $||\b{y}||_2 = 1$. Changing the dummy variable $\b{y}$ to $\b{x}$ gives the Eq. (\ref{equation_rayleigh_ritz_quotient_optimization}).
The $(b)$ notices $\b{x}^\top\b{x} = 1$ because of the constraint $||\b{x}||_2=1$.

Note that the constraint in Eq. (\ref{equation_rayleigh_ritz_quotient_optimization}) can be equal to any constant which is proved similarly. Moreover, note that the value of constant in the constraint is not important because it will be removed after taking derivative from the Lagrangian in optimization \cite{boyd2004convex}.

The \textit{generalized Rayleigh-Ritz quotient} or \textit{generalized Rayleigh quotient} is defined as \cite{parlett1998symmetric,ghojogh2019eigenvalue}:
\begin{align}\label{equation_generalized_rayleigh_ritz_quotient}
\mathbb{R} \ni R(\b{A}, \b{B}; \b{x}) := \frac{\b{x}^\top \b{A}\, \b{x}}{\b{x}^\top \b{B}\, \b{x}},
\end{align}
where $\b{A}$ and $\b{B}$ are symmetric matrices and $\b{x}$ is a non-zero vector:
\begin{align}
\b{A} = \b{A}^\top, ~~ \b{B} = \b{B}^\top, ~~ \b{x} \neq \b{0}.
\end{align}
If the symmetric $\b{B}$ is positive definite:
\begin{align}
\b{B} \succ 0,
\end{align}
it has a Cholesky decomposition:
\begin{align}\label{equation_generalized_rayleigh_ritz_quotient_B_Cholesky}
\b{B} = \b{C} \b{C}^\top,
\end{align}
where $\b{C}$ is a lower triangular matrix. 
In case $\b{B} \succ 0$, the generalized Rayleigh-Ritz quotient can be converted to a Rayleigh-Ritz quotient:
\begin{align}
R(\b{A}, \b{B}; \b{x}) = R(\b{D}, \b{C}^\top \b{x}),
\end{align}
where:
\begin{align}\label{equation_generalized_rayleigh_ritz_quotient_D}
\b{D} := \b{C}^{-1} \b{A} \b{C}^{-\top}.
\end{align}
The proof is:
\begin{align*}
\text{RHS} &= R(\b{D}, \b{C}^\top \b{x}) \overset{(\ref{equation_rayleigh_ritz_quotient})}{=} \frac{(\b{C}^\top \b{x})^\top \b{D}\, (\b{C}^\top \b{x})}{(\b{C}^\top \b{x})^\top (\b{C}^\top \b{x})} \\
&\overset{(\ref{equation_generalized_rayleigh_ritz_quotient_D})}{=} \frac{\b{x}^\top \b{C} \b{C}^{-1} \b{A} (\b{C} \b{C}^{-1})^\top \b{x}}{\b{x}^\top (\b{C} \b{C}^\top) \b{x}} \overset{(a)}{=} \frac{\b{x}^\top \b{A}\, \b{x}}{\b{x}^\top \b{B}\, \b{x}} \\
&\overset{(\ref{equation_generalized_rayleigh_ritz_quotient})}{=} R(\b{A}, \b{B}; \b{x}) = \text{LHS}, ~~~~~ \text{Q.E.D.},
\end{align*}
where RHS and LHS are short for right and left hand sides and $(a)$ is because of Eq. (\ref{equation_generalized_rayleigh_ritz_quotient_B_Cholesky}) and $\b{C}\b{C}^{-1} = \b{I}$ because $\b{C}$ is a square matrix.

Similarly, one of the properties of the generalized Rayleigh-Ritz quotient is:
\begin{align}\label{equation_generalized_rayleigh_ritz_quotient_scaling}
R(\b{A}, \b{B}; c\b{x}) = R(\b{A}, \b{B}; \b{x}),
\end{align}
where $c$ is a scalar. The proof is that:
\begin{align*}
R(\b{A}, \b{B}; c\b{x}) &= \frac{(c\b{x})^\top \b{A}\, c\b{x}}{(c\b{x})^\top \b{B}\, c\b{x}} \overset{(a)}{=} \frac{c\b{x}^\top \b{A}\, c\b{x}}{c\b{x}^\top \b{B}\, c\b{x}} \\
&\overset{(b)}{=} \frac{c^2}{c^2} \times \frac{\b{x}^\top \b{A}\, \b{x}}{\b{x}^\top \b{B}\, \b{x}} \overset{(\ref{equation_generalized_rayleigh_ritz_quotient})}{=} R(\b{A}, \b{B}; \b{x}),
\end{align*}
where $(a)$ and $(b)$ are because $c$ is a scalar.

Because of the Eq. (\ref{equation_generalized_rayleigh_ritz_quotient_scaling}), the optimization of the generalized Rayleigh-Ritz quotient has an equivalent:
\begin{align}
&\underset{\b{x}}{\text{minimize/maximize}} ~~ R(\b{A}, \b{B}; \b{x}) \equiv \nonumber \\
&\begin{aligned}\label{equation_generalized_rayleigh_ritz_quotient_optimization}
& \underset{\b{x}}{\text{minimize/maximize}}
& & \b{x}^\top \b{A}\, \b{x} \\
& \text{subject to}
& & \b{x}^\top \b{B}\, \b{x} = 1,
\end{aligned}
\end{align}
for a similar reason that we provided for the Rayleigh-Ritz quotient. the constraint can be equal to any constant because in the derivative of Lagrangian, the constant will be dropped.

\bibliography{References}
\bibliographystyle{icml2016}

\end{document}